\documentclass{article}
\usepackage{iclr2026_conference,times}

\usepackage{amsmath,amsfonts,bm}

\def\eqref#1{equation~\ref{#1}}

\def\1{\bm{1}}

\DeclareMathAlphabet{\mathsfit}{\encodingdefault}{\sfdefault}{m}{sl}
\SetMathAlphabet{\mathsfit}{bold}{\encodingdefault}{\sfdefault}{bx}{n}

\newcommand{\E}{\mathbb{E}}

\newcommand{\R}{\mathbb{R}}

\usepackage{hyperref}
\usepackage{url}

\usepackage{graphicx}
\usepackage{booktabs}
\usepackage{multirow}
\usepackage{wrapfig}
\usepackage{amsmath}
\usepackage{amssymb}
\usepackage{algorithm}
\usepackage{algpseudocode}
\usepackage[accsupp]{axessibility}
\usepackage{pifont}
\usepackage{xcolor}

\AtBeginDocument{%
  \setlength{\abovedisplayskip}{6pt plus 2pt minus 2pt}%
  \setlength{\belowdisplayskip}{6pt plus 2pt minus 2pt}%
  \setlength{\abovedisplayshortskip}{3pt plus 1pt minus 1pt}%
  \setlength{\belowdisplayshortskip}{4pt plus 1pt minus 1pt}%
}
\makeatletter
\def\section{\@startsection{section}{1}{\z@}{-1.6ex plus -0.5ex minus -.2ex}%
  {1.1ex plus 0.2ex minus 0.2ex}{\large\sc\raggedright}}
\def\subsection{\@startsection{subsection}{2}{\z@}{-1.4ex plus -0.5ex minus -.2ex}%
  {0.6ex plus .2ex}{\normalsize\sc\raggedright}}
\def\paragraph{\@startsection{paragraph}{4}{\z@}{1.2ex plus 0.4ex minus .2ex}%
  {-1em}{\normalsize\bf}}
\makeatother

\newcommand{\tabsqueezefactor}{0.85}
\newcommand{\figsqueezefactor}{0.85}
\makeatletter
\newcommand{\cap@scalefont}[1]{%
  \dimen@=\f@size pt\relax
  \dimen@=#1\dimen@
  \dimen@ii=\baselineskip
  \dimen@ii=#1\dimen@ii
  \fontsize{\strip@pt\dimen@}{\strip@pt\dimen@ii}\selectfont
}
\newcommand{\cap@scale}{}
\AtBeginDocument{%
  \let\cap@origmakecaption\@makecaption
  \long\def\@makecaption#1#2{%
    \begingroup\cap@scale\cap@origmakecaption{#1}{#2}\endgroup}%
}
\newcommand{\tabsqueeze}[1][\tabsqueezefactor]{%
  \cap@scalefont{#1}%
  \setlength{\tabcolsep}{#1\tabcolsep}%
  \def\cap@scale{\cap@scalefont{#1}}%
}
\newcommand{\figsqueeze}[1][\figsqueezefactor]{%
  \def\cap@scale{\cap@scalefont{#1}}%
}
\makeatother

\newcommand{\latent}{z}
\newcommand{\Zsp}{\mathcal{Z}}
\newcommand{\proprio}{\text{proprio}}
\newcommand{\proj}{\mathrm{proj}_{\Zsp}}
\newcommand{\zrew}{\latent_{\mathrm{rew}}}

\title{\textsc{CrossBFM:} Distilling a Shared Latent Behavior Space Across Humanoid Embodiments}

\author{Tan-Dzung Do$^{1*}$,\ Tuan Dat Phuong$^{1,2*}$\footnotemark[4]\hspace{4pt},\ Nico Bohlinger$^{3}$,\ Cuc T. Trinh$^{1}$,\ Siwei Ju$^{3}$, \\
{\bf Vien Anh Ngo$^{1,4}$,\ Jan Peters$^{3,5,6}$,\ Xinchao Wang$^{2\dagger\text{\ding{41}}}$,\ An T. Le$^{1,3,4\dagger}$} \\[2pt]
{\normalfont $^{1}$VinRobotics \quad $^{2}$National University of Singapore \quad $^{3}$TU Darmstadt} \\
{\normalfont $^{4}$VinUniversity \quad $^{5}$DFKI \quad $^{6}$Hessian AI} \\[2pt]
{\normalfont\small $^{*}$Equal contribution \quad $^{\dagger}$Equal advising \quad \ding{41}\,Corresponding author}
}

\iclrfinalcopy
\begin{document}
{\renewcommand{\thefootnote}{\fnsymbol{footnote}}%
 \footnotetext[4]{Work partially done during internship at VinRobotics}}
\maketitle

\begin{figure}[h]
  \centering
  \includegraphics[width=0.9\textwidth]{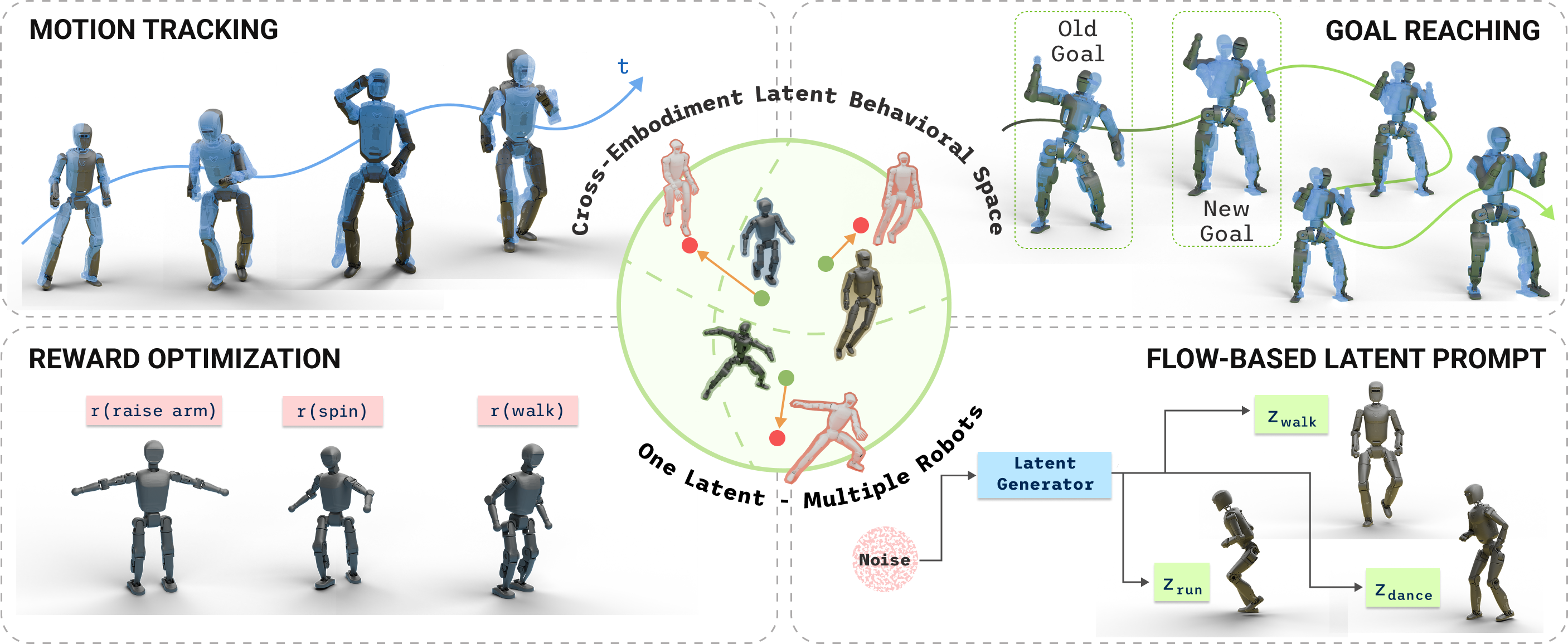}
  \caption{\textsc{CrossBFM} leverages a unified encoder to distill a frozen source BFM latent onto new robots. The distilled latent can process various prompts, including the three original modes -- motion tracking, goal reaching, and reward optimization -- and also a new mode for flow-based generated latents.}
  \label{fig:teaser}
  \vspace{-\intextsep}
\end{figure}

\begin{abstract}
Behavior Foundation Models (BFMs) give humanoids a promptable policy over a latent behavior space, enabling one single vector to represent a motion to imitate, a pose to reach, or a reward to maximize. Forward-Backward representations successfully produce such spaces, but at the cost of hundreds of GPU-hours for a single robot. Moreover, when the training process is repeated for a second robot, it produces a second space unrelated to the first, resulting in embodiment-specific latents that do not unify or transfer. We address these problems with \textsc{CrossBFM}, treating the latent space as the transferable asset for various embodiments. As retargeting provides frame-level cross-embodiment correspondence, we propose a unified encoder architecture with no robot-specific parameters for distilling the behavior space to address all training embodiments simultaneously in less than a GPU-hour. Following this encoder, latent-conditioned trackers turn the distilled latent into whole-body control in a conventional PPO training manner in just 10 more GPU-hours. On three distilled humanoids, all three prompting modes transfer: motion tracking with latent-conditioned policy losing only $0.025$ rad to its joint-conditioned counterpart, smooth goal reaching between poses with no falls, and reward optimization for all $41$ reward prompts. Our experiments further reveal that 1) regressing the encoder on a quarter of the motion corpus costs only $5\%$ of tracking performance and 2) training the encoder on a subset of robots and evaluating on an unseen one recovers up to $89\%$ of the tracking performance of seen robots, demonstrating cross-embodiment generalization to morphologically similar robots. We also verify the pipeline on real robots across all three prompting modes and with flow-based generated latents. Project website: \url{https://dotandung.github.io/crossbfm/}
\end{abstract}

\section{Introduction}
\label{sec:intro}
Humanoid whole-body control via Reinforcement Learning (RL) policies has witnessed remarkable progress with robots performing a wide range of motions, ranging from dancing or parkour to full loco-manipulation in challenging settings~\citep{yang2026omniretarget, liao2026beyondmimic, wu2026perceptive, luo2026sonic}. Commanding these policies typically requires a sequence of actions in joint space~\citep{ze2025twist, ze2025twist2, liao2026beyondmimic, do2026compliantwbc}, that were retargeted specifically for the robot to mimic. Besides joint-conditioned control, whole-body control, through Forward--Backward (FB)
representations~\citep{touati2021fb,touati2023fb,tirinzoni2025fbcpr}, established a compelling
command interface in the latent space for humanoid robots via motion latent features. In this paradigm, a policy $\pi(\cdot \mid \latent)$, conditioned on a latent
vector $\latent$, that is drawn from a trained behavioral space $\Zsp \subseteq \R^d$, can be prompted at test time to track a
motion, reach a goal pose, or maximize a commanded reward, all without any retraining.
BFM-Zero~\citep{li2025bfmzero} brought this idea from simulated characters~\citep{tirinzoni2025fbcpr} onto real robots, followed by frameworks such as UFO~\citep{roboparty2026ufo}, which further improve the training
pipeline. Nevertheless, one training run still costs more than 100 GPU hours for any robot. 

While the pretrained BFM can handle various prompt types, it is always trained for one specific robot. For a new robot, the training has to run from scratch, which in turn produces a new latent space that does not related to one from a previous robot. Although both latent spaces are geometrically similar,
the coordinates at a given point correspond to different behaviors. Since every downstream task depends on this latent behavior space, the resulting cross-embodiment transfer becomes as problematic as the doubled training cost. Previous cross-embodiment learning works introduce alternative approaches to expand BFM without retraining. Some~\citep{bohlinger2024urma,patel2025getzero,ai2025scaling} propose to condition one policy on an embodiment description and train with multiple embodiments simultaneously, which generalizes to unseen embodiments. While these methods demonstrated cross-embodiment transfer, they are typically constrained to a single
task (e.g. velocity-tracking locomotion or in-hand rotation) with their shared representation encoding the morphology rather than task-conditioned behavior. These frameworks lack representations encoding ``the reward I want maximized'' or ``the pose I
want reached'' compared to BFM, which directly encodes the action value for each latent coordinate of the learned space (\emph{value-functional}).

\begin{figure}[t]
  \centering
  \includegraphics[width=\textwidth]{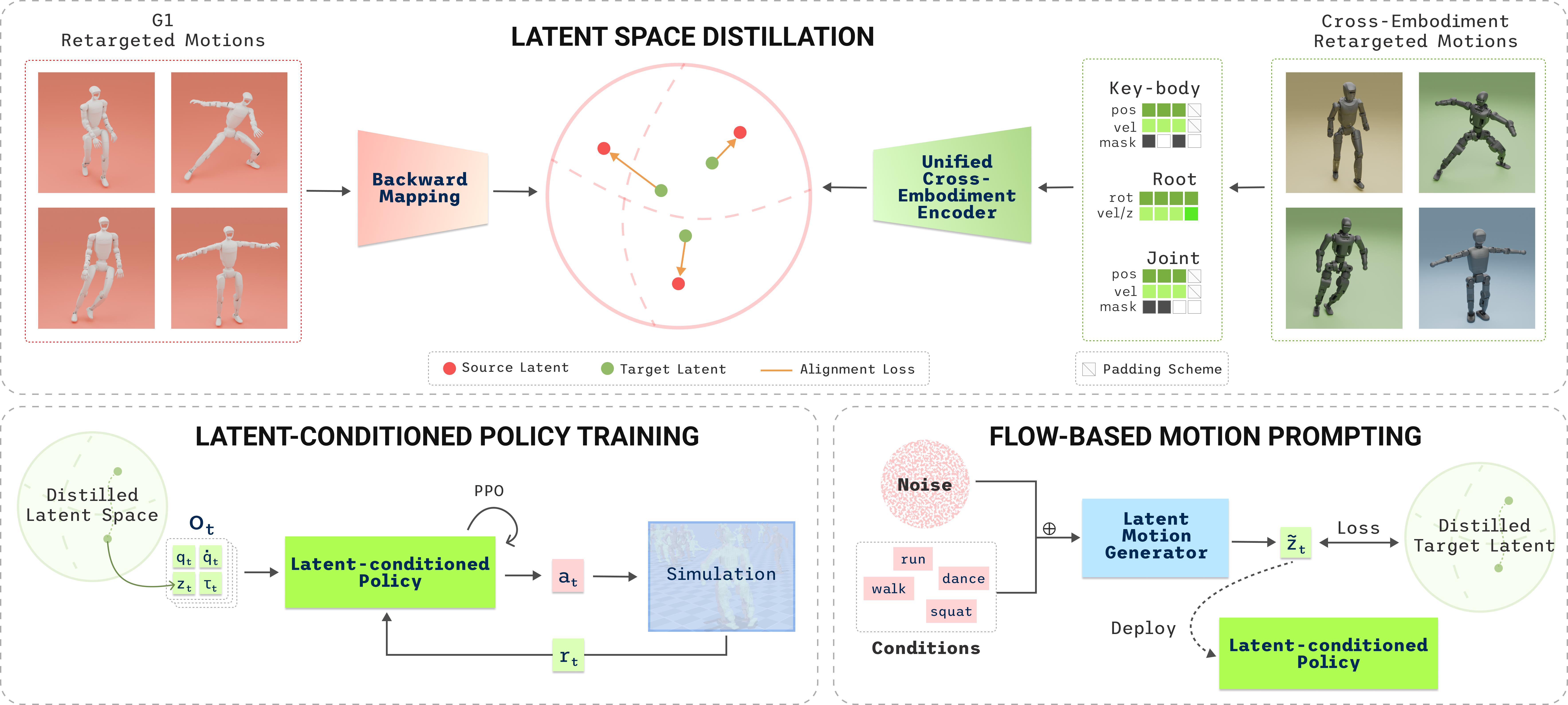}
  \vspace{-\intextsep}
  \caption{\textsc{CrossBFM} overview. \emph{Top:} a unified encoder encoder mapping all new robots' proprioception onto the frozen source latent using frame-level
  correspondences from retargeted data. \emph{Bottom:} a $\latent$-conditioned policy and a flow-based
  latent generator on top of the distilled latent space.}
  \label{fig:pipeline}
\end{figure}

\textbf{Our approach.} We propose \textsc{CrossBFM} (Figure~\ref{fig:pipeline}) to tackle both the repeated training and latent space mismatch problems. First, we \emph{freeze} a BFM pretrained for a specific robot, viewing its FB latent as a
fixed \emph{behavioral coordinate system}, and distill that coordinate system onto new embodiments. By leveraging retargeted motions for different robots along the same timeline, we can directly establish cross-embodiment correspondences for all the training robots, which reduces learning the backward map for the target robot to supervised regression. After this correspondence is established, we introduce a unified encoder to map all robot configurations into a shared latent space in only one training run. Our encoder is a
fixed-width view of the robot proprioception with $33$ joint and $8$ key-body indices, into
which any robot's degrees of freedom are placed. Indices that do not exist for a specific robot are
masked-out, such that the input dimensionality is
identical across embodiments and the architecture becomes universal for a wide range of humanoids. Finally, we train \emph{latent-conditioned trackers} that receive latents from the shared space, enabling whole-body control across various embodiments. While the
source model spent on the order of $10^2$ GPU-hours of online unsupervised RL --- with a replay
buffer, a discriminator, and domain randomization --- to obtain a backward map for one robot, our full pipeline requires just around 10 hours on an RTX4090.

\textbf{Contributions.}
\begin{itemize}
  \item \textbf{A unified encoder for BFM distillation.} We introduce a robot-independent encoder architecture allowing for multi-robot distillation in one training at
  a comparable accuracy of separate per-robot encoders,
  while producing more consistent latents across robots and generalizing to unseen embodiments of a similar morphology. (Section~\ref{sec:slots}).
  \item \textbf{Retargeting as a correspondence oracle.} We show how latent transfer across
  humanoid embodiments is equivalent to supervised regression from a frozen source BFM latent,
  thanks to frame-level correspondence across retargeted datasets.
  This removes the simulator, RL training, and adversarial objective from the target robot side (Section~\ref{sec:stage1}).
  \item \textbf{Latent-condition whole-body control.} We evaluate our method on three distinct humanoids with trackers conditioned on the distilled latent $[\latent \mid \proprio]$, achieving comparable results with their joint-conditioned counterparts while covering all three prompting modes of BFM-Zero. We extend the latent space by introducing a flow-based latent generator that takes in behavior modes and generates the corresponding latents. We also verify our pipeline on real robots, demonstrating its transfer to real hardware (Section~\ref{sec:exp}).
\end{itemize}
\vspace{-1em}
\section{Related Work}
\label{sec:related}
\vspace{-0.5em}

\paragraph{Behavior foundation models and unsupervised RL for humanoid control.}
Most humanoid whole-body controllers are trained as motion \emph{trackers} with an on-policy RL
algorithm such as PPO~\citep{schulman2017ppo} optimizing an explicit imitation reward for a
retargeted reference trajectory~\citep{luo2023phc,cheng2024expressive,he2024omnih2o,ze2025twist2}.
This family of policies requires a per-frame reference in joint space, which is challenging to obtain for complex tasks without a high-level planner. Character animation instead learns a
\emph{latent} skill space from unlabeled motion and conditions a policy on
the trained latent space~\citep{peng2022ase,tessler2023calm}, gaining reusability at the price of a latent whose
semantics are only implicitly defined. Unsupervised RL~\citep{gregor2016vic,eysenbach2019diayn,pathak2017curiosity} and in particular the
Forward--Backward family~\citep{touati2021fb,touati2023fb} give the latent an explicit meaning as a task descriptor. In this setting, a reward, a goal, or a
demonstration map to a vector in the latent space that the downstream policy conditions on. Following this line of work, FB-CPR~\citep{tirinzoni2025fbcpr}
made this practical for high-dimensional humanoids by regularizing the unsupervised policy toward an
unlabeled motion dataset with a latent-conditional discriminator while BFM-Zero~\citep{li2025bfmzero}
extends this further to a real robot through domain
randomization and safety-oriented reward shaping. UFO~\citep{roboparty2026ufo} rebuilt the
infrastructure for speed and generality, cutting FB pre-training by five times to over 100 hours on a consumer GPU and showing that
other unsupervised objectives, e.g. temporal-distance representations~\citep{bae2024tldr}, can enhance the latent consistency. All of these works produce a behavior space \emph{per robot}, i.e. unrelated to other robots. \textsc{CrossBFM} directly wires the pretrained space of one robot to other embodiments without reruning the costly FB training.

\vspace{-0.5em}
\paragraph{Cross-embodiment learning.}
Training a single policy across many robots requires architectures and training paradigms that can either condition on or abstract over embodiment differences.
Previous works propose Graph Neural Networks (GNNs) to directly use the kinematic structure of robots as part of the network~\citep{wang2018, huang2020one}, and more recent works use attention-based architectures that leverage body parts as tokens~\citep{gupta2022metamorph,sferrazza2025body,patel2025getzero,ai2025scaling,bohlinger2026shape} or infer the embodiment from long interaction histories~\citep{liu2025locoformer, li2026rapid}.
These approaches achieve generalization to unseen embodiments but focus on a single task with the shared component being the embodiment encoder rather than behavior space. A second family directly establishes correspondence without policy conditioning, aligning state spaces of two policies
with optimal transport~\citep{fickinger2022gwil} or graph matching~\citep{le2025kinematics}.
This approach is often more expensive, while \textsc{CrossBFM} direcly leverages retargeted datasets as cross-embodiment correspondences. A third and more recent
family learns unified cross-embodiment latent
spaces~\citep{yan2026unified,kim2026phasor,unit2026,motif2026}. These are closest to our work by intuition,
but their latents are learned from scratch and are descriptive rather than \emph{value-functional}, thus they
do not come with a closed form that directly turns a reward into a latent. \textsc{CrossBFM} differs on both aspects. Our latent space is not learned but \emph{inherited} from a frozen FB model, and it therefore
retains the FB prompting semantics on the new embodiments it is distilled onto.

\vspace{-0.5em}
\paragraph{Retargeting and distillation.}
Motion retargeting maps a human mocap trajectory onto a robot's kinematics conditioned on embodiment-specific characteristics such as joint limits or key body constraints. Modern retargeting pipelines are accurate and fast enough to be run both offline over large dataset as well as responsive enough for real-time teleoperation~\citep{ze2026gmr,ze2025twist2,yang2026omniretarget}. In this work, we leverage retargeted data as ground-truth correspondence for cross-embodiment transfer. We formulate our training objective to be a form of knowledge
distillation~\citep{hinton2015distillation} with the teacher (source BFM) and the students (new robots) observing
\emph{embodiment-specific} retargeted motions aligned to each other by the shared timeline.

\section{Preliminaries}
\label{sec:prelim}

\subsection{Unsupervised RL and successor measures}
We consider a reward-free discounted Markov decision process
$\mathcal{M}=(S,A,P,\mu,\gamma)$, with state space $S$, action space $A$, transition kernel
$P(\mathrm{d}s'\mid s,a)$ taking over all possible states $s'$ over an infinitesimal region around it $ds'$, initial-state distribution $\mu$ and discount $\gamma\in(0,1)$. Because no
reward is given at training time, the object an unsupervised RL agent can learn is the
\emph{dynamics} of its own policies. For a policy $\pi$, the \emph{successor
measure}~\citep{dayan1993sr,blier2021successor} records where that policy goes after infinite steps $t$,
\begin{equation}
  M^{\pi}(X\mid s,a) \;:=\; \sum_{t\ge 0}\gamma^{t}\,\Pr\bigl(s_{t+1}\in X \,\big|\, s,a,\pi\bigr),
  \qquad X\subseteq S.
  \label{eq:succ}
\end{equation}
This representation is particularly useful as it factorizes the value function into successor probability multiplied by the reward gained at that state. Specifically, for \emph{any} reward $r:S\to\R$,
\begin{equation}
  Q^{\pi}_{r}(s,a) \;=\; \int_{S} M^{\pi}(\mathrm{d}s'\mid s,a)\, r(s') .
  \label{eq:qsucc}
\end{equation}
Equation~\ref{eq:qsucc} separates ``how the policy moves'' from ``how much reward the policy gains'', which can be used to evaluate any reward at test time without any reward-driven finetuning.

\subsection{Forward--Backward representations}
FB representations~\citep{touati2021fb,touati2023fb} realize the unsupervised RL objective by taking a finite-rank
approximation of the successor measure. Given a state distribution $\rho$, one learns a
\emph{forward} map $F:S\times A\times\Zsp \to \R^{d}$ and a \emph{backward} map $B:S\to\R^{d}$, along
with a latent-conditioned policy $\pi_\latent$, such that
\begin{equation}
  M^{\pi_\latent}(\mathrm{d}s' \mid s,a) \;\simeq\; F(s,a,\latent)^{\!\top} B(s')\,\rho(\mathrm{d}s'),
  \qquad
  \pi_\latent(s) \;=\; \arg\max_{a} F(s,a,\latent)^{\!\top}\latent ,
  \label{eq:fb}
\end{equation}
where $\Zsp\subseteq\R^{d}$ is conventionally the sphere of radius $\sqrt{d}$, and $F,B$ are trained
to minimize the temporal-difference residual of the measure-valued Bellman
equation~\citep{touati2021fb,tirinzoni2025fbcpr}. Substituting Equation~\ref{eq:fb} into
Equation~\ref{eq:qsucc} gives the closed-form of the latent inference for any reward $r$,
\begin{equation}
  Q^{\pi_{\latent_r}}_{r}(s,a) \;=\; F(s,a,\latent_r)^{\!\top}\latent_r ,
  \qquad
  \latent_r \;=\; \E_{s\sim\rho}\bigl[\,B(s)\,r(s)\,\bigr].
  \label{eq:zrew}
\end{equation}
Consequently, the backward map $B$ converts a reward function into the latent whose policy maximizes it
in \emph{closed form}.

\subsection{The three prompting modes}
\label{sec:modes}
With $F$, $B$ and $\pi_\latent$ trained by the objectives in ~\citet{tirinzoni2025fbcpr}, a BFM can answer three kinds of prompts at test time without
any retraining or planning:
\begin{itemize}
  \item \textbf{Motion tracking}: given a reference motion $\tau=(s_1,\dots,s_n)$, the latent at
  time $t$ is a look-ahead embedding of the reference given by $\latent_t = \proj\bigl(\sum_{t'=t}^{t+H} B(s_{t'})\bigr)$.
  \item \textbf{Goal reaching}: given a target state $s_g$, then $\latent_g = \proj\bigl(B(s_g)\bigr)$.
  \item \textbf{Reward optimization}: given samples $\{(s_i,r_i)\}_{i=1}^{M}$ with $s_i\sim\rho$,
  use the empirical form of Equation~\ref{eq:zrew}, $\zrew = \proj\bigl(\sum_i \omega_i\, r_i\, B(s_i)\bigr)$.
\end{itemize}
The backward map $B$ enables the smooth conversion from desired behaviors to corresponding latents, which then drive the policy. This is the structural reason why our method focuses on the backward map for direct latent distillation rather than both the forward and backward maps. If we can, for a new embodiment, produce latents in the frozen source's coordinate system,
we directly inherit all three prompting modes at once via the closed-form computation of $B$.

\section{\textsc{CrossBFM}}
\label{sec:pipeline}

\subsection{Setup and notation}
We freeze a source BFM trained on the Unitree G1~\citep{li2025bfmzero} and focus only on its backward map $B_S$.
We denote $d=256$ and $\Zsp$ is the sphere of radius $\sqrt{d}$, where $\proj$ is the radial
projection onto it. For every frame $t$ of a reference motion, we compute the source latent
\begin{equation}
  \latent^\star_t \;=\; \proj\bigl(B_S(o^{G1}_t)\bigr),
  \label{eq:zstar}
\end{equation}
which is the ground-truth latent label retained from the source model. Each clip is retargeted independently
onto every robot $X$ on the \emph{same} timeline with the same frequency, so the target frame $o^X_t$ and the
source frame $o^{G1}_t$ correspond to the same motion of the same behavior. We then attach $\latent^\star_t$ to their corresponding retargeted data as ground-truth latent. The correspondence that cross-domain imitation normally has to learn is therby handled by the
retarget algorithm. Consequently, the target-side learning problem becomes the regression
$E_X: o^X_{1:N} \mapsto \hat\latent_{1:N} \in \Zsp$ that maps a \emph{new} robot's proprioception onto
a latent that a \emph{source} robot produces.

\subsection{The unified encoder architecture}
\label{sec:slots}
Different humanoids may vary in the number of actuated joints, the joint order, and their body composition. A per-robot encoder
absorbs these differences explicitly by having a different input configuration, but then requires multiple networks to be trained without a shared representation. This design also hinders generalization to new robots with different morphologies. We instead define a fixed-width, robot-independent input, covering joints, key bodies and root configuration, then map each robot's
retargeted data into this unified input for the encoder.

Our encoder focuses on two main morphological components. First, \emph{8 key-body indices}, comprising
$\{\text{left},\text{right}\}\times\{\text{knee},\text{foot},\text{elbow},\text{arm tip}\}$. Here, the root
is deliberately excluded from the key-body set , since key-body positions are expressed relative to it thus the root row would be
simply zero. Second, \emph{33 canonical joint indices} include $12$
leg, $3$ waist (yaw / roll / pitch), $16$ arm and $2$ head joints. This vocabulary intentionally contains
indices that \emph{no} robot in our experiments have (such as waist roll and pitch or head joints), such
that the encoder can be universally applied for a wide range of robots. Missing indices are
\emph{masked} with a binary flag and zero-padded, such that the network can tell ``this joint is at zero'' from ``this joint does not exist.'' We additionally include a \emph{robot pelvis proprioception} that carries the root's global height, orientation and linear velocity, which serves as the global anchor for other key bodies. More details are provided in Appendix~\ref{app:slots}.

\subsection{Stage 1: backward map regression}
\label{sec:stage1}

The unified encoder takes as input a short window of length $T$ of the proprioception in a reference motion of one robot and returns a
latent for every frame in it. First, we use a linear layer to map each frame to the encoder embedding width $H$, followed by a positional encoding to embed the frame order. We add pre-LayerNorm transformer
blocks~\citep{vaswani2017transformer,xiong2020preln} to allow frames to attend to one another, which improve the temporal consistency and prevent mode collapse. Finally, we use a linear layer to turn each frame into the shape of the frozen BFM latent,
rescaled to the sphere of radius $\sqrt{d}$
(more in Appendix~\ref{app:enc-arch}). The latent projection is given by:
\begin{equation}
  \hat\latent_t \;=\; \proj\bigl(E_X(o^X_{t-T+1:t})\bigr), \qquad
  \proj(v) \;=\; \sqrt{d}\; v / \max(\|v\|,\varepsilon).
  \label{eq:enc}
\end{equation}
We introduce \emph{causal} attention to our encoder that applies an upper-triangular attention mask so that
frame $t$ attends only to frames $\le t$. While the bidirectional version sees roughly one second
of the \emph{future reference} and remains acceptable for offline motion processing, it is unsuitable for real-world control. The objective is a frame-wise cosine regression onto the
frozen source latent:
\begin{equation}
  \mathcal{L}_{\rm align} \;=\; \frac{1}{N}\sum_{t}\Bigl(1-\cos\bigl(\hat\latent_t,\latent^\star_t\bigr)\Bigr).
  \label{eq:align}
\end{equation}
The cosine distance serves as a proximity measure for the cross-embodiment representation on the latent space. With this simple architecture and objective, the encoder learns to map the target robot's proprioception to the source latent space in just under an hour on a single consumer GPU, against the $\sim\!10^2$ GPU-hours that
are needed to train $B_S$. Our encoder has no robot-specific
weights and each robot configuration enters only through the fixed index maps in
Section~\ref{sec:slots}. Appendix~\ref{app:encoder} compares our unified encoder architecture against three
separate per-robot encoders.

\subsection{Stage 2: latent-conditioned policies}
With the distilled latent space, we then train a tracking policy $\pi_X$ for each robot with PPO~\citep{schulman2017ppo}. These trackers are latent-conditioned, meaning that they only
take as input the behavior latent and proprioception. The proprioception includes base angular velocity, IMU roll and
pitch, joint positions and velocities, and the previous action. In this setting, there is no explicit reference trajectory in
the actor's observation apart from the latent behavior. We keep the critic asymmetric with the privileged reference observations available in simulation. The privileged critic keeps the value estimation accurate without leaking unavailable information into the policy. Because the latent 
$\latent$ is shared across embodiments, we can leverage a generative model to approximate this space to command \emph{every} robot. In this work, we use a single rectified-flow
model~\citep{liu2023rectifiedflow,lipman2023flowmatching} trained over latent chunks, so that a behavior can be given by a wider range of prompts (text for example), rather than restricted to robot-specific
joint-space references. More about the latent generator in Appendix~\ref{app:flow}.

\subsection{Deployment: prompting without a target-side FB model}
\label{sec:reward}
Two of the three prompting modes of Section~\ref{sec:modes} immediately transfer after the encoder training. Tracking reference is
$\hat\latent_t = E_X(o^X_{t-T+1:t})$ on a retargeted reference while goal reaching's is $E_X$ on a predefined window
ending at the goal pose. However, reward optimization requires post-processing on the target side, as Equation~\ref{eq:zrew} requires two components from the frozen source BFM. While the backward map $B_S$ can be directly substituted with the trained encoder $E_X$, the distribution $\rho$ of visited states collected during RL training, over which the expectation is taken, is not available on the target robot that never ran unsupervised RL. To address this, we approximate the distribution $\rho$ for the distilled robots using their retargeted datasets. This ensures that the reward-weighted projection selects states that are feasible for the target robot, rather than relying on the source's distribution, which may contain poses that are not reachable on the target side. For the reward function, to address the variation in joint configurations, we keep only the key-body kinematics (root height and velocity, uprightness, angular velocity, arm-tip and knee heights) as reward features and discard terms on the joint configuration. Specifically, we first leverage the trained encoder to convert proprioception into latents and compute the reward features for each pose in the cached dataset over selected features. We then compute a reward-weighted average over all cached latents as the final reference latent for the policy to track:
\begin{equation}
  \underbrace{Z_i = E_X\bigl(o^X_{i-T+1:i}\bigr)}_{\text{encode once}}, \qquad
  \underbrace{r_i = r\bigl(\phi(o^X_i);\theta^{(X)}\bigr)}_{\text{score cached features}}, \qquad
  \underbrace{\zrew = \proj\Bigl(\textstyle\sum_i \omega_i\, r_i\, Z_i\Bigr)}_{\text{project}},
  \label{eq:zrew_target}
\end{equation}
with $\omega = \mathrm{softmax}(\tau\, r_{1:M})$, $M$ cached frames of the retargeted dataset, and a temperature $\tau$. Another problem remains, which is that the source-robot reward often refers to
absolute configurations, such as a specific hand height or reach distance, that a differently proportioned
robot cannot attain. We propose to leverage the keybody distribution of each robot over the retargeted dataset, so that keybody positions of different robots would lie in a similar pose percentile within the shared dataset. For example, the ``raise arm" task of a medium robot remarked by $1.0$\,m hand height would lie in the $p$ percentile of its feature distribution. With the $p$ percentile as anchor for the other robots' feature distribution, it may convert to $0.820$\,m on a smaller robot and $1.157$\,m on a larger robot. Given $\Phi_S,\Phi_r$ being the CDFs of key-body features for source and target robots, respectively, the transport is therefore specified by: 
\begin{equation}
  \theta^{(r)} \;=\; \Phi_r^{-1}\bigl(\Phi_S(\theta)\bigr).
  \label{eq:transport}
\end{equation}
 With this alignment, we can analytically compute the reward formulation for all new robots simultaneously without having to hand-craft thresholds for each robot. Details in Appendix~\ref{app:transport}.

\section{Experiments}
\label{sec:exp}

\paragraph{Setup.} We distill a frozen BFM pretrained on the Unitree G1 robot~\citep{li2025bfmzero} onto three target robots (Inhouse M3, Booster T1, Fourier N1)
that differ in their number of joints, links and their topology. Following the pretrained BFM, we use the LAFAN dataset~\citep{harvey2020lafan} for our experiments. More about the task settings in Appendix~\ref{app:tasks}. We ask 3 questions:
\begin{itemize}
\item How well do the three original prompting modes transfer to distilled robots in the training set, both in simulation and in the real world?
\item Can the unified encoder generalize to unseen motion within the trained latent space, both functionally for motion tracking and geometrically for representation reconstruction? Since the source space is well-structured, how much data do we need to distill this space?
\item Can the encoder generalize to unseen robots?
\end{itemize}

\begin{table}[t]
  \centering
  \small
  \setlength{\tabcolsep}{4pt}
  \caption{Latent-conditioned vs. joint-conditioned trackerby joint mean absolute error.}
  \label{tab:track}
  \begin{tabular}{llcccccc}
    \toprule
    \multirow{2}{*}{Robot} & \multirow{2}{*}{DoF\,/\,bodies}
      & \multicolumn{3}{c}{Tracking MAE (rad) $\downarrow$}
      & \multicolumn{2}{c}{Goal MAE (rad) $\downarrow$} \\
    \cmidrule(lr){3-5} \cmidrule(lr){6-7}
     & & Joint \ (TWIST2) & Latent (ours) & $\Delta_{\rm cond}$ & Latent (ours) & $\Delta_{\rm goal}$ \\
    \midrule
    M3 & $27/30$ & $0.1777 \pm 0.0033$ & $0.2024 \pm 0.0067$ & $+0.0247$ & $0.2345 \pm 0.0011$ & $+0.0321$ \\
    T1 & $23/24$ & $0.1844 \pm 0.0043$ & $0.1901 \pm 0.0032$ & $+0.0057$ & $0.1967 \pm 0.0012$ & $+0.0066$ \\
    N1 & $23/29$ & $0.1362 \pm 0.0032$ & $0.1391 \pm 0.0038$ & $+0.0029$ & $0.1512 \pm 0.0023$ & $+0.0121$ \\
    \bottomrule
  \end{tabular}
\end{table}

\subsection{Transferring three original prompting modes}
\label{sec:exp:track}

\begin{wraptable}{r}{0.46\textwidth}
  \centering
  \small
  \setlength{\tabcolsep}{4pt}
  \caption{Reward optimization prompt results by normalized returns. \emph{Source-side} computes $\zrew$ on the G1 with the true backward map $B_S$; \emph{target-side (ours)} computes it via
  Equation~\ref{eq:zrew_target}.}
  \label{tab:reward}
  \begin{tabular}{lccc}
    \toprule
    Robot & Source & Target (ours) & Gain \\
    \midrule
    M3 & $0.145$ & $\mathbf{0.508}$ & $3.50\times$ \\
    T1 & $0.228$ & $\mathbf{0.261}$ & $1.14\times$ \\
    N1 & $0.177$ & $\mathbf{0.523}$ & $2.95\times$ \\
    \midrule
    Mean & $0.183$ & $\mathbf{0.431}$ & $2.35\times$ \\
    \bottomrule
  \end{tabular}
\end{wraptable}

\paragraph{Tracking and Reaching.} We compare latent-conditioned policies against joint-conditioned counterparts trained by
TWIST2~\citep{ze2025twist2} under the same training settings (Table~\ref{tab:track}). The
joint-conditioned policies are naturally the upper bound for the tracking task as at every control step, they have access to exactly which
joint angles to track. In contrast, the latent-conditioned policies are only given a single latent corresponding to the desired behavior and must infer the pose from it.

Nevertheless, across all three robots, the gap (latent $-$ joint) stays within $0.025$\,rad and the latent-conditioned policies successfully follow the reference motions. Replacing a
dense per-frame reference with one latent vector therefore marginally hurts the performance while allowing for
 smooth transition between poses, the property that the joint-conditioned policies do not have. In fact, for the joint-conditioned policies, commanding a discontinuous jump in joint targets produces large torques and
early termination~\citep{roboparty2026ufo}. For this reason, goal reaching is only reported for the latent-conditioned policies,
where interpolating in $\Zsp$ results in smooth transitions at a cost of at most $0.032$\,rad compared to the tracking performance. The latent-conditioned policies also express prompted behaviors generated by the flow-based
latent generator within the distilled space, including walking, running and dancing (see our website). Together, this suggests that
a sufficiently fine-grained latent space with more complex primitive behaviors could let an operator command a humanoid through
\emph{behavioral intent} rather than joint targets, which is a substantially more natural interface for complex
long-horizon loco-manipulation.

\paragraph{Reward optimization.}Table~\ref{tab:reward} compares the latent produced by using the state buffer collected during RL training with the true backward map $B_S$ of the source model versus the latent produced by the cached retargeted dataset with our percentile reward threshold using the feature transformation. Computing $\zrew$ on
the source robot with the true backward map $B_S$ is $3.5\times$ and $3.0\times$ worse on the M3 and N1 compared to ours. The reason is that
the reward-weighted projection selects \emph{which states are worth visiting}, which depends
on which states the executing robot can actually reach. Thus, a latent, which is optimal for the G1's reachable set, is
not optimal for other robots.
Together with tracking and goal reaching, this covers all three BFM prompting
modes~\citep{touati2023fb,tirinzoni2025fbcpr} on new robots, demonstrating
\emph{functional} property for a shared behavior space.

\subsection{Latent inference on unseen motions}
\label{sec:exp:generalize}
 
\begin{wraptable}{r}{0.60\textwidth}
  \centering
  \small
  \setlength{\tabcolsep}{4pt}
  \caption{Closed-loop key-body error \texttt{mpjpe\_local}$\downarrow$ ($\times10^{-2}$\,m) for the tracking task using the oracle latent
  $\latent^\star$ versus the encoder's inference, split by training/validation set.}
  \label{tab:generalize}
  \begin{tabular}{lcccccc}
    \toprule
    \multirow{2}{*}{Encoder split} & \multicolumn{2}{c}{M3} & \multicolumn{2}{c}{T1} & \multicolumn{2}{c}{N1} \\
    \cmidrule(lr){2-3} \cmidrule(lr){4-5} \cmidrule(lr){6-7}
     & oracle & infer. & oracle & infer. & oracle & infer. \\
    \midrule
    Training ($30$)   & $8.14$ & $8.02$ & $5.70$ & $5.43$ & $5.58$ & $5.57$ \\
    Validation ($10$) & $6.40$ & $7.07$ & $5.26$ & $5.60$ & $5.46$ & $5.60$ \\
    \bottomrule
  \end{tabular}
\end{wraptable}

We evaluate how well the unified encoder infers latent behaviors across the training and validation motions, both functionally for the tracking task and geometrically for latent alignment.

\textbf{Functionality.}  Table~\ref{tab:generalize} demonstrates the performance of latent trackers conditioned on oracle and inferred latents in . On the encoder's training clips, the
predicted latent is marginally better than the oracle latent $\latent^\star$. On validation motions, the latent costs $1.4$--$6.7$\,mm of key-body error while maintaining good balance for the whole trajectories. Since all eight behaviors of the LAFAN dataset appear in
the training set, we conclude that the encoder is able to generalize to new \emph{clips} of similar behaviors.

\begin{figure}[t]
  \centering
  \includegraphics[width=0.95\textwidth]{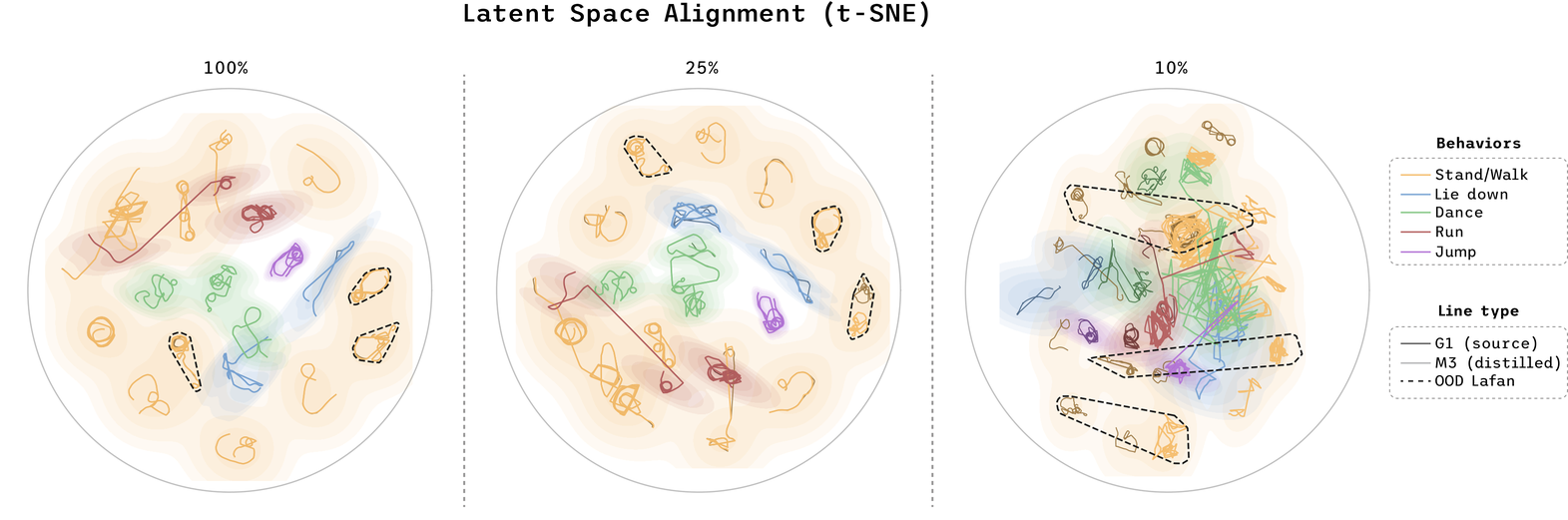}
  \caption{t-SNE projection of source (G1, dark) and distilled (M3, bright) latent trajectories, colored by behavior type. Dashed outlines represent validation clips (\emph{unseen} clips of similar behaviors).}
  \label{fig:tsne}
\end{figure}

\textbf{Geometrical alignment.} Figure~\ref{fig:tsne}a projects source
and distilled latent trajectories into 2D, showing how distilled
trajectories \emph{overlay} the source trajectories \emph{almost identically}. This means that our encoder enforces latents to live in the source's coordinate system while forming clear behavioral clusters. The validation clips (dashed) also fall clearly \emph{inside} the cluster of
their own behavior type rather than besides the manifold or in the gaps between the clusters. For example, an unseen walking clip
is not mapped somewhere new but rather into the region of $\Zsp$, that the walking cluster already occupies. As a result,
the tracker trained on that region can adapt to validation latents well (Table~\ref{tab:generalize}).

\subsection{Data scaling: a quarter of the corpus suffices}
\label{sec:exp:scaling}

\begin{wrapfigure}{r}{0.44\textwidth}
  \centering
  \includegraphics[width=0.41\textwidth]{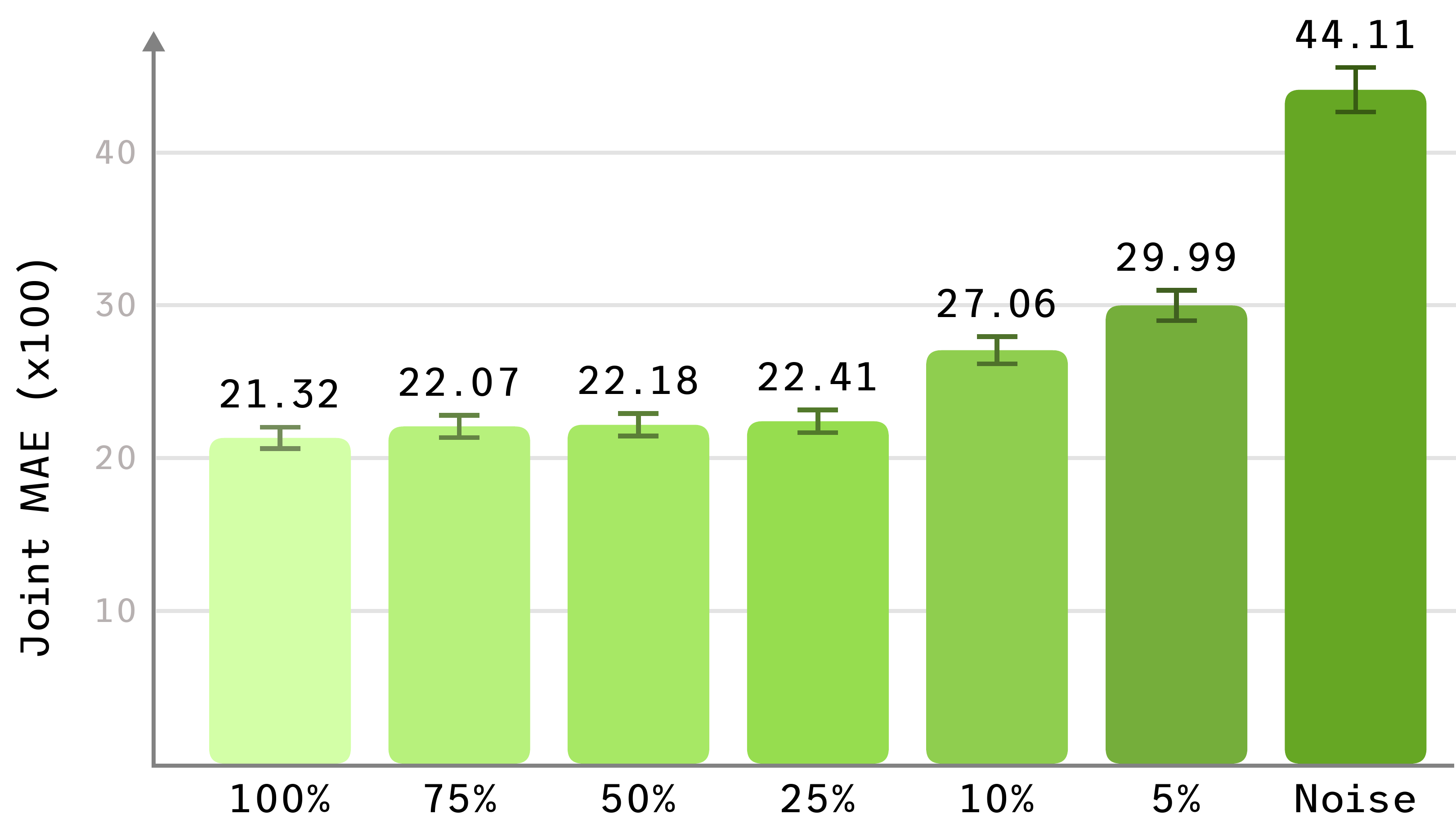}
  \caption{Data scaling for the M3 robot, evaluated by closed-loop joint MAE as we shrink the distillation corpus from
  $100\%$ to $5\%$ to only noise.}
  \label{fig:scaling}
\end{wrapfigure}

We shrink the distillation dataset up to $95\%$ by randomly selecting $K=\sum ^{40}_1 K_i$ non-overlapping segments of $1.5$--$3$\,s for all 40 motions. The total length of these $K$ segments add up to $T_K \in(5, 75)$ percent of the LAFAN dataset. We choose continuous segments rather than isolated frames, such that local temporal structure is
preserved and all the $40$ behaviors remain available. These segments are then used to train unified encoders with shrunk dataset. After training, we feed the latent inferred by these encoders to a tracker trained on 100\% of the data to see how well the shrunk encoders can reconstruct \emph{behaviors}. Figure~\ref{fig:scaling} shows how the joint MAE only rises
from $0.2132$, at $100\%$ of the corpus, to $0.2241$, at $25\%$, demonstrating a $5\%$ degradation for a four times less data. The tracking performance only starts to break down at $10\%$ ($0.2706$) 
against a random-latent control at $0.4411$. The geometry degrades gradually similar to the tracking performance. From Figure~\ref{fig:tsne}b, at $25\%$, the distilled trajectories still overlay the source with the behavior classes
clearly separated, even for the validation clips, unlike the tangled structure at $10\%$ data. Hence, a quarter of the corpus not only retains comparable tracking performance but also reconstructs the \emph{geometry} of the source's behavior space.

\subsection{Generalize to unseen robots}
\label{sec:exp:loro}

\vspace{-5pt}
\begin{table}[t]
  \centering
  \small
  \setlength{\tabcolsep}{5pt}
  \caption{Generalization to unseen robots. \emph{2-R E.} represents the encoder trained on two robots and evaluated on the third,
  with fixed pretrained trackers. \emph{3-R E.}\ is the unified encoders
  trained on all 3 robots.}
  \label{tab:loro}
  \begin{tabular}{lcccccc}
    \toprule
    \multirow{2}{*}{Unseen robot}
      & \multicolumn{2}{c}{Validation cosine $\uparrow$}
      & \multicolumn{3}{c}{Closed-loop joint MAE (rad) $\downarrow$}
      & \multirow{2}{*}{Signal gap closed $\uparrow$} \\
    \cmidrule(lr){2-3} \cmidrule(lr){4-6}
     & 2-R E. & 3-R E. & 2-R E. & 3-R E. & random z \\
    \midrule
    M3 & $0.6251 \pm 0.0196$ & $0.8658$ & $0.1829 \pm 0.0054$ & $0.1645$ & $0.3371$ & $\mathbf{89.3\% \pm 3.5}$\\
    N1 & $0.5861 \pm 0.0100$ & $0.8640$ & $0.2237 \pm 0.0048$ & $0.1370$ & $0.3320$ & $\mathbf{55.5\%\pm 3.2}$\\
    T1 & $0.2533 \pm 0.0168$ & $0.8300$ & $0.2886 \pm 0.0138$ & $0.1679$ & $0.3063$ & $\mathbf{12.8\%\pm 10.3}$\\
    \bottomrule
  \end{tabular}
\end{table}

Table~\ref{tab:loro} reports the generalizability of the unified encoder to unseen robots by training it on two robots and evaluating on the third one. Apart from cosine similarity and tracking performance, we also report
$\text{signal gap closed}=(\text{random}-\text{pred})/(\text{random}-\text{oracle})$, represeting how meaningful the latent signal is to the tracker, with the upper bound being the seen robot and the lower bound being pure noise. 
The cross-embodiment transfer works best for morphologically similar robots. Specifically, an encoder that never saw M3 recovers $89.3\%$ of the
tracking performance of the encoder trained on all 3 robots, while for unseen T1, it recovers only $12.8\%$.

\textbf{Cosine distance does not fully capture policy performance.} With the cosine distance alone, the latents of unseen robots lie far away form the ground-truth, illustrated by $0.625$ cosine similarity for the best case with M3. However, this encoder still generates latents that recover $89.3\%$ performance of trackers trained on the oracles' latent. While cosine similarity directly answers how close $\hat\latent$ is to $\latent^\star$, it does not ask
whether the policy reproduces the correct behavior. Our latent-conditioned policy is robust to some latent mismatch, provided that the inferred latents fall into the behavior cluster that the policy can effectively utilize, even when the cosine distance appears large.

\textbf{Cosine distance relates to morphological proximity.} In our experiments, M3 and N1 are a close pair
both in terms of joint configuration and body shape, while T1 is smaller and significantly different in its morphology. From Table~\ref{tab:loro}, when the unseen
robot still has a close relative in training, cosine similarity lands at $0.59$--$0.63$. In contrast, when the unseen robot is the T1 with no close relative, the latent similarity drops to $0.25$. This result suggests that \emph{given a significantly large set of embodiments where every new robot has a close
relative, generalization to new robots then becomes interpolation rather than extrapolation}. Our finding aligns well with the conclusion from ~\citep{bohlinger2025multi}, in which
locomotion policies trained with thousands of randomized embodiments demonstrated zero-shot transfer to unseen
robots.

\subsection{Real-world deployment}
\label{sec:exp:real}

\begin{figure}[h]
  \centering
  \includegraphics[width=\textwidth]{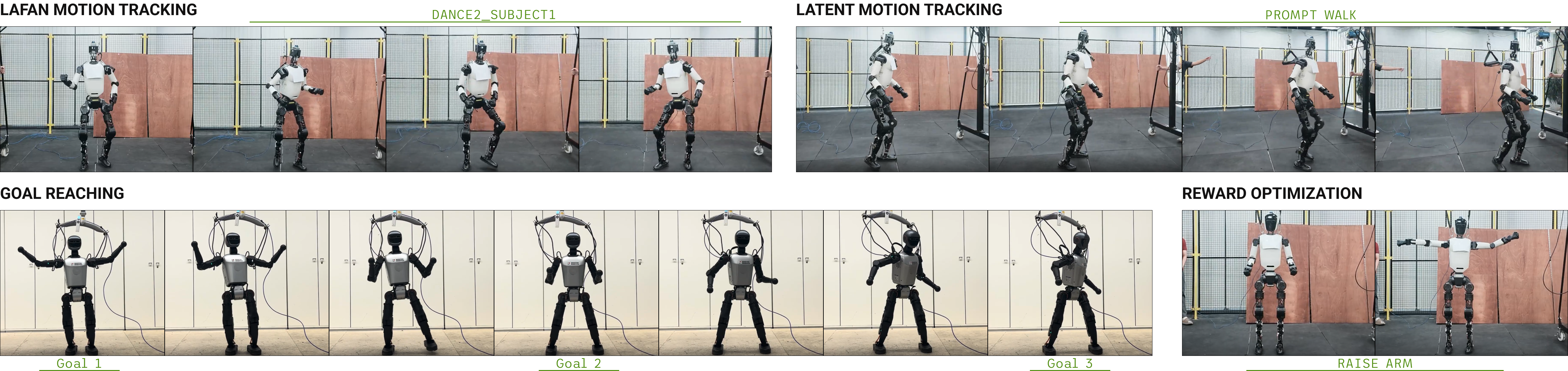}
  \caption{Real-world rollouts of T1 and M3 across all 4 prompt modes. Videos in our \href{https://dotandung.github.io/crossbfm}{website}.}
  \label{fig:real}
\end{figure}

We verify the performance of our latent-conditioned policy with its joint-conditioned counterpart in the real world on the M3 and T1 robots. Table~\ref{tab:real} shows the quantitative results on the M3 with all 3 prompt modes, with joint-conditioned policy staying slightly ahead of our latent-condition policy, similar as in simulation. For the goal reaching and reward optimization tasks, the latent-conditioned policy also performs comparable with the results in simulation (Table~\ref{tab:track}) with some small degradation due to sim-to-real gap.

\begin{table}[t]
  \centering
  \small
  \setlength{\tabcolsep}{4pt}
  \caption{Sim2real results on 3 prompt modes: motion tracking (T), goal reaching (G) and reward optimization (R). Metrics: T and G
  are joint MAE in radians, R is normalized return.}
  \label{tab:real}
  \begin{tabular}{llcc}
    \toprule
    Mode & Prompt & Real & Sim \\
    \midrule
    T (joint ref.)   & $5$ clips & $0.1985$ & $0.1777$ \\
    T (latent, ours) & $5$ clips & $0.2206$ & $0.2024$ \\
    G                & $9$ poses  & $0.2675$ & $0.2345$ \\
    R                & $3$ tasks  & $0.562$ & $0.508$ \\
    \bottomrule
  \end{tabular}
\end{table}

\section{Conclusion}
\label{sec:conclusion}
\textsc{CrossBFM} transfers a behavior space across humanoid embodiments by treating retargeted data of different robots using the same dataset as cross-embodiment correspondence and distilling the latents of a frozen source BFM through supervised
learning. We propose a unified encoder with no per-robot
parameters that convert proprioception into latent behavior in one training run under a GPU hour. Our unified encoder also enables generalization to unseen motion of similar behaviors and to unseen robot of similar morphology. The tracking policies conditioned on distilled latents achieve tracking performances comparable to their joint-conditioned counterpart while allowing for smooth
goal reaching by latent interpolation, and zero-shot reward prompting across $41$ tasks. We also demonstrate, that with our pipeline, using only one-quarter of the dataset is enough to effectively distill the frozen latent space onto a new robot, retaining both the geometrical alignment and functionality. Future works may focus on the scaling law across two main axes, including richer training data for a more fine-grain behavior space and a larger robot training set for better generalization to unseen robots. Additionally, with our shared latent space, prompting with text instructions to address long-horizon loco-manipulation tasks is also a promising direction.

\section*{Acknowledgment}

This project was supported in part by the National Science Centre Poland in the Weave programme UMO2021/43/I/ST6/02711, the German Science Foundation (DFG) under grant number PE 2315/17-1, the German Federal Ministry of Education and Research (BMBF) and the Hessian Ministry of Science and Research, Art and Culture (HMWK).

\newpage

\bibliography{iclr2026_conference}
\bibliographystyle{iclr2026_conference}

\newpage
\appendix

\section{Implementation and Training Details}
\label{app:impl}

\subsection{Robot platforms}
\label{app:robots}
Table~\ref{tab:robots} describes the four embodiments in our experiment. The source is a frozen BFM on the Unitree
G1, from which we retain only the backward map $B_S$. The three targets differ from G1, and from
each other, in DoF count, body count, joint names and joint order.
\begin{table}[h]
  \centering
  \small
  \setlength{\tabcolsep}{4pt}
  \caption{The four humanoid configurations.}
  \label{tab:robots}
  \begin{tabular}{llcccl}
    \toprule
    Robot & Role & DoF & Bodies & Index filled & Morphology \\
    \midrule
    Unitree G1 & source & $29$ & $32$  & ---     & reference topology; $3$-DoF waist, $3$-DoF wrists \\
    M3         & target & $27$ & $30$  & $27/33$ & nearest to the source; waist yaw only \\
    Booster T1 & target & $23$ & $24$ & $21/33$ & wrist-less, $4$-DoF arms, $2$-DoF head \\
    Fourier N1 & target & $23$ & $29$ & $23/33$ & one wrist DoF per arm; waist yaw only \\
    \bottomrule
  \end{tabular}
\end{table}

\subsection{The encoder input}
\label{app:slots}
In Table~\ref{tab:slots}, we provide the full view of the encoder input, which is previously described in Section~\ref{sec:slots}. In this setting, every robot presents the same $163$ dimensions
regardless of how many degrees of freedom it has, and the presence masks tell the encoder whether the current entry is present or missing.
\begin{table}[h]
  \centering
  \caption{The unified cross-embodiment encoder canonical input.}
  \label{tab:slots}
  \begin{tabular}{llc}
    \toprule
    Range & Contents & Dims \\
    \midrule
    $[0{:}24)$    & key-body positions, root-relative, $/\,\textsc{scale}$ & $8\times3$ \\
    $[24{:}48)$   & key-body velocities (finite difference $\times$ fps)   & $8\times3$ \\
    $[48{:}56)$   & key-body presence mask                                 & $8$ \\
    $[56{:}57)$   & root height $/\,\textsc{scale}$                        & $1$ \\
    $[57{:}61)$   & root rotation quaternion                               & $4$ \\
    $[61{:}64)$   & root linear velocity $/\,\textsc{scale}$               & $3$ \\
    $[64{:}97)$   & joint positions $q$, canonical slot order              & $33$ \\
    $[97{:}130)$  & joint velocities $\dot q$                              & $33$ \\
    $[130{:}163)$ & joint presence mask                                    & $33$ \\
    \midrule
    \multicolumn{2}{l}{\textbf{Total}} & $\mathbf{163}$ \\
    \bottomrule
  \end{tabular}
\end{table}

\subsection{Encoder architecture}
\label{app:enc-arch}

\begin{figure}[h]
  \centering
  \includegraphics[width=0.5\textwidth]{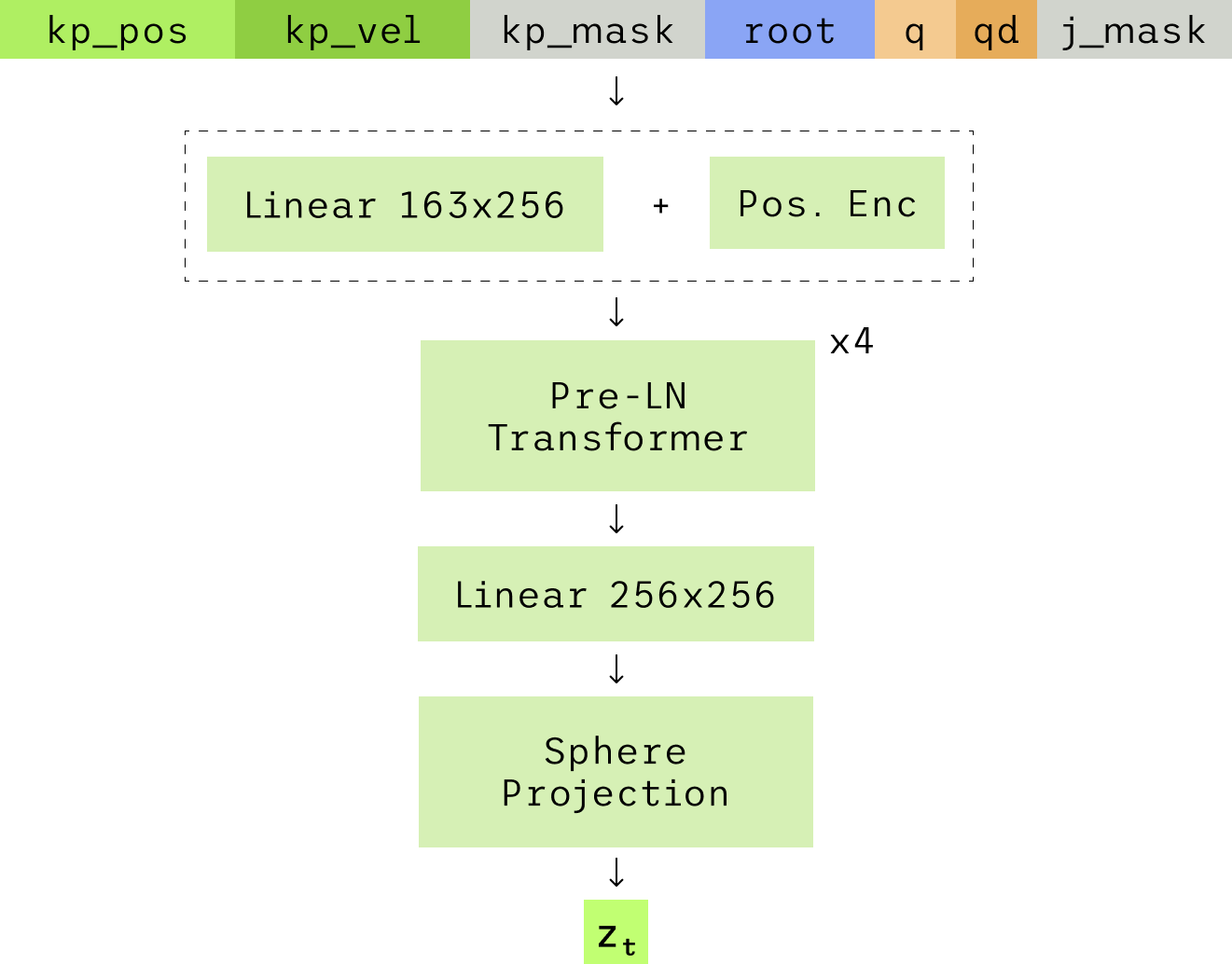}
  \figsqueeze
  \caption{Unified encoder architecture.}
  \label{fig:app:encoder}
\end{figure}

The encoder is a sequence-to-sequence transformer that maps a window of canonical proprioceptive frames to one
latent per frame. We provide its architecture in Table~\ref{tab:enc-arch}. Most of that
configuration is conventional with two crucial design choices. The first is
\emph{pre-LayerNorm}, which turned out to be necessary to prevent mode collapse.  When we used post-LayerNorm instead, the model
collapsed to one constant latent, and the cosine objective cannot recover because every prediction then points in the same direction. The second is the \emph{output
projection onto the sphere}, which mirrors the source model's own \texttt{project\_z} and therefore
prevents the encoder from emitting a latent that falls out of the manifold of the frozen tracker. We make this projection part of the network rather than a post-processing step, so that
gradients flow through it during training.

Three more conventions make the encoder universal, which are \emph{fixed}
data conventions rather than learned parameters, so none of them adds per-robot weights to the model.
\begin{enumerate}
  \item \textbf{Name maps.} The key-body indices follow the same correspondence that the trackers were
  trained against. The joint index order
  follows the joint ordering that the retargeting scripts use.
  \item \textbf{Sign canonicalization.} T1 measures waist yaw with the opposite sign convention to
  G1, M3 and N1. We therefore provide an additional index for this joint rotation convention.
  \item \textbf{Robot scale normalization.} We divide all lengths by $\textsc{scale}$, which is the median
  root height of that robot over its own corpus ($0.7956$\,m for M3, $0.6743$\,m for N1 and
  $0.6552$\,m for T1). This normalization is what lets a $0.65$\,m robot and a $0.80$\,m robot present
  geometrically comparable inputs to the same encoder.
\end{enumerate}

With these conventions, adding new robots only requires a
name map and a scale constant, which ask for only configuration file rather than additional training or architecture-wise modification.

\begin{table}[h]
  \centering
  \small
  \caption{Encoder configuration.}
  \label{tab:enc-arch}
  \begin{tabular}{@{}l p{0.60\textwidth}@{}}
    \toprule
    Component & Value \\
    \midrule
    \multicolumn{2}{l}{\emph{Input}} \\
    Input width                & $163$ (Table~\ref{tab:slots}) \\
    Window length $T$          & $64$ frames $=2.13$\,s at $30$\,fps \\
    Input projection           & $\mathrm{Linear}(163 \to H)$ \\
    Positional encoding        & sinusoidal, max length $1024$, non-persistent buffer \\
    \midrule
    \multicolumn{2}{l}{\emph{Trunk}} \\
    Layers                     & $4 \times$ pre-LayerNorm transformer encoder layers \\
    Attention heads            & $8$ \\
    Feed-forward width         & $4H$ \\
    Activation                 & GELU \\
    Dropout                    & $0$ \\
    Attention mask             & bidirectional (default); upper-triangular in the causal variant \\
    \midrule
    \multicolumn{2}{l}{\emph{Head}} \\
    Output head                & $\mathrm{LayerNorm}(H) \to \mathrm{Linear}(H \to 256)$ \\
    Output projection          & $\latent \leftarrow \sqrt{256}\, v / \max(\|v\|, \varepsilon)$, $\varepsilon=10^{-6}$ \\
    Output rate                & one latent per frame (not per window) \\
    \midrule
    \multicolumn{2}{l}{\emph{Width and parameter count}} \\
    Unified encoder            & $H=256$, $3.27$M parameters total (one checkpoint, all robots) \\
    Per-robot encoders         & $H=512$, $12.8$M parameters each, $38.5$M total \\
    Per-robot conditioning     & none; robot identity enters only through the fixed slot maps \\
    \bottomrule
  \end{tabular}
\end{table}

\subsection{Encoder training}
\label{app:enc-train}
Table~\ref{tab:enc-train} lists the training protocol used for
encoder ablation experiments in this paper. For data sampling, \texttt{MultiRobot\-Window\-Dataset} draws a single window index and then returns \emph{the
same window} for all $R$ robots, and it supervises all $R$ copies coupled with a shared ground-truth
$\latent^\star$. Every optimizer step therefore presents $R$ morphological views of one motion against
a shared label.
\begin{table}[h]
  \centering
  \small
  \caption{Encoder training protocol across experiments.}
  \label{tab:enc-train}
  \begin{tabular}{@{}l p{0.60\textwidth}@{}}
    \toprule
    Parameter & Value \\
    \midrule
    Objective            & $\mathcal{L}=\frac{1}{N}\sum_t\bigl(1-\cos(\hat\latent_t,\latent^\star_t)\bigr)$ \\
    Optimizer            & AdamW \\
    Learning rate        & $2\times10^{-4}$ \\
    Weight decay         & $10^{-4}$ \\
    Gradient clipping    & global norm $1.0$ \\
    Epochs               & $400$ \\
    Batch size (unified) & $22$ windows $\times\ 3$ robots $=66$ \\
    Batch size (per-robot) & $64$ windows \\
    Window length        & $64$ frames \\
    Training data        & LAFAN, $30$ clips, $195{,}746$ frames per robot \\
    Validation data      & $10$ clips (\texttt{walk:3,*:1} quota) \\
    Evaluation           & validation cosine every $10$ epochs \\
    Wall-clock           & under one hour on a single consumer GPU \\
    \bottomrule
  \end{tabular}
\end{table}

\subsection{Tracker architecture and training}
\label{app:tracker}
Each embodiment receives its own tracking policy trained with
PPO~\citep{schulman2017ppo} in mjlab. The
two conditioning policies (latent-conditioned and joint-conditioned) of Table~\ref{tab:track} run in the \emph{same} environment with two different configurations. First, we swap the motion command for a variant that additionally carries
the frozen G1 latent. Second, we remove the five reference-derived terms
(\texttt{motion\_root\_vel\_xy\_b}, \texttt{motion\_root\_z}, \texttt{motion\_root\_roll\_pitch},
\texttt{motion\_root\_yaw\_ang\_vel\_b}, \texttt{motion\_joint\_pos}) from the actor and replace them
with $\latent$. Everything else is shared between the two variants: the rewards, the terminations, the
randomization, the PPO budget, and even the motion \texttt{.pkl} files, which carry $\latent$ as one
extra field alongside the standard tracking reference. We leave the critic with privileged future reference untouched in both
variants, which exists only in simulation and is discarded
at deployment.

\begin{table}[h]
  \centering
  \small
  \caption{Tracker configuration and PPO hyperparameters.}
  \label{tab:tracker}
  \begin{tabular}{@{}l p{0.62\textwidth}@{}}
    \toprule
    Parameter & Value \\
    \midrule
    \multicolumn{2}{@{}l}{\emph{Environment}} \\
    Simulator            & mjlab (MuJoCo), flat terrain \\
    Physics rate         & $200$\,Hz (timestep $0.005$\,s) \\
    Control rate         & $50$\,Hz (decimation $4$) \\
    Episode length       & $10$\,s $=500$ control steps \\
    Parallel environments & $4096$ \\
    Action space         & joint position targets for all actuated DoF, per-robot action scale, applied as an offset from the default pose \\
    Phase sampling       & uniform over the clip \\
    \midrule
    \multicolumn{2}{@{}l}{\emph{Observations}} \\
    Actor, current       & $[\latent^{256} \mid \proprio]$: base angular velocity ($\times0.25$), IMU roll/pitch, $q$, $\dot q$ ($\times0.05$, ankle DoF masked to $0$), previous action \\
    Actor, history       & proprioception only, $11$ steps, encoded to a $64$-dim latent \\
    Joint-conditioned variant & replaces $\latent^{256}$ with the reference terms listed above \\
    Critic               & privileged future reference sequence $+$ current proprioception $+$ privileged extras (base linear velocity, root pose, key-body positions, foot contact, and the sampled randomization parameters) \\
    \midrule
    \multicolumn{2}{@{}l}{\emph{Networks}} \\
    Actor / critic trunk & MLP $(512, 256, 128)$, ELU, running observation normalization \\
    Action distribution  & Gaussian, scalar state-independent std, $\sigma_0=1.0$ \\
    History encoder      & $64$-dim latent; motion/command encoder $64$-dim \\
    \midrule
    \multicolumn{2}{@{}l}{\emph{PPO}} \\
    Steps per environment per update & $24$ \\
    Iterations           & $30{,}000$ ($\approx\!2.9\times10^{9}$ environment steps) \\
    Learning rate        & $10^{-3}$, adaptive schedule with target ${\rm KL}=0.01$ \\
    Discount $\gamma$ / GAE $\lambda$ & $0.99$ / $0.95$ \\
    Clip range           & $0.2$, with clipped value loss \\
    Value / entropy coefficient & $1.0$ / $0.005$ \\
    Learning epochs / minibatches & $5$ / $4$ \\
    Max gradient norm    & $1.0$ \\
    Seed                 & $0$ for the A/B; $3$ seeds for the numbers in Table~\ref{tab:track} \\
    Wall-clock           & under ten hours per embodiment on one RTX\,4090 \\
    \bottomrule
  \end{tabular}
\end{table}

\begin{table}[h]
  \centering
  \small
  \caption{Tracker reward terms.}
  \label{tab:reg}
  \setlength{\tabcolsep}{3pt}
  \begin{tabular}{@{}llc@{\hskip 1.2em}llc@{}}
    \toprule
    \multicolumn{3}{@{}l}{\emph{Tracking}} & \multicolumn{3}{l}{\emph{Regularization}} \\
    \cmidrule(r){1-3} \cmidrule(l){4-6}
    Term & & Weight & Term & & Weight \\
    \midrule
    joint position      & & $2.0$ & alive              & & $+0.5$ \\
    joint velocity      & & $0.2$ & feet air time & & $1.0$ \ ($5.0$) \\
    key-body position   & & $2.0$ & action rate   & & $-0.03$ \ ($-0.1$) \\
    key-body position (global) & & $2.0$ & joint acceleration & & $-2.5\!\times\!10^{-8}$ \ ($-5\!\times\!10^{-8}$) \\
    root height         & & $1.0$ & ankle joint acc. & & $-2\!\times\!10^{-7}$ \ ($-1\!\times\!10^{-7}$) \\
    root rotation       & & $1.0$ & joint limit        & & $-10.0$ \\
    root linear velocity & & $1.0$ & DoF position limits & & $-5.0$ \\
    root angular velocity & & $1.0$ & self collision    & & $-10.0$ \\
                        & &        & DoF torque limits  & & $-1.0$ \\
                        & &        & feet stumble       & & $-1.25$ \\
                        & &        & feet slip          & & $-0.1$ \\
                        & &        & body angular vel.\ $xy$ & & $-0.01$ \\
                        & &        & joint velocity     & & $-1\!\times\!10^{-4}$ \\
                        & &        & ankle joint velocity & & $-2\!\times\!10^{-4}$ \\
                        & &        & feet contact force & & $-5\!\times\!10^{-4}$ \\
    \bottomrule
  \end{tabular}
\end{table}

\subsection{Domain randomization and observation noise}
\label{app:dr}
Both conditioning variants train under identical randomization, listed in Table~\ref{tab:dr}. Domain randomization is disabled during the evaluation rollouts.

\begin{table}[h]
  \centering
  \small
  \caption{Domain randomization.}
  \label{tab:dr}
  \setlength{\tabcolsep}{4pt}
  \begin{tabular}{@{}lll p{0.42\textwidth}@{}}
    \toprule
    Event & Mode & Operation & Range \\
    \midrule
    Pelvis added mass       & startup  & add   & $[-3.0, 3.0]$\,kg \\
    Pelvis COM offset       & startup  & add   & $x\!:\![-0.025,0.025]$, $y,z\!:\![-0.05,0.05]$\,m \\
    Motor strength ($k_p$, $k_d$) & startup & scale & $[0.8, 1.2]$ \\
    Foot friction           & startup  & set   & $[0.3, 1.2]$\\
    Joint encoder bias      & startup  & add   & $[-0.01, 0.01]$\,rad \\
    Actuator delay          & startup  & ---   & $0$--$4$ physics steps, per-environment phase \\
    Push (base velocity)    & interval every $1$--$3$\,s & set & $v_{x,y}\!:\!\pm0.5$, $v_z\!:\!\pm0.2$\,m/s; $\omega_{r,p}\!:\!\pm0.52$, $\omega_y\!:\!\pm0.78$\,rad/s \\
    \midrule
    \multicolumn{4}{@{}l}{\emph{Additive uniform observation noise (actor only)}} \\
    Base angular velocity   & \multicolumn{2}{l}{} & $\pm0.1$ \\
    IMU roll / pitch        & \multicolumn{2}{l}{} & $\pm0.1$ \\
    Joint position $q$      & \multicolumn{2}{l}{} & $\pm0.01$ \\
    Joint velocity $\dot q$ & \multicolumn{2}{l}{} & $\pm0.1$ \\
    \bottomrule
  \end{tabular}
\end{table}

\section{Tasks and Metrics}
\label{app:tasks}

\subsection{Metrics}
\label{app:metrics}

\paragraph{Policy-free metrics.} \emph{Alignment} is the validation cosine between the encoder's prediction and
the frozen source latent, $\cos(\hat\latent_t, \latent^\star_t)$, averaged over the validation frames.
\emph{Agreement} is the cosine $\cos(\hat\latent_{r,t}, \hat\latent_{r',t})$ between two robots'
predictions on the same validation frame.

\paragraph{Closed-loop metrics with policy.} For motion tracking and goal reaching task, we report the following metrics. \emph{Joint MAE} is the mean absolute deviation, in radians, between the
executed joint positions and the retargeted reference, averaged over the episode and over the joints.
\emph{Key-body error} $\texttt{mpjpe\_local}$ is the mean Euclidean distance, in metres, between the
executed and the reference key-body positions, both expressed in the root frame.

\subsection{Reward optimization task suite}
\label{app:rewards}
We evaluate reward prompting on the $41$ task rewards designed for G1 robot~\citep{li2025bfmzero} (Table~\ref{tab:rewardsuite}). Each reward is a
function of the state, normalized to $[0,1]$. For every task we infer a single $\zrew$
from the new robot's cached buffer (Equation~\ref{eq:zrew_target}) and then hold that latent constant for
$300$ control steps, which is one full episode. We score only the settled second half of the episode after the robot reached a stable pose.

\begin{table}[h]
  \centering
  \small
  \caption{The $41$ source task rewards by group.}
  \label{tab:rewardsuite}
  \begin{tabular}{@{}llp{0.60\textwidth}@{}}
    \toprule
    Category & Count & Tasks \\
    \midrule
    Standing    & $2$  & \texttt{move-ego-0-0}, \texttt{move-ego-low0.5-0-0} \\
    Locomotion  & $8$  & \texttt{move-ego-0-0.7}, \texttt{move-ego-90-0.7}, \texttt{move-ego--90-0.7}, \texttt{move-ego-low0.6-0-0.7}, \texttt{move-ego-0-0.3}, \texttt{move-ego-90-0.3}, \texttt{move-ego-180-0.3}, \texttt{move-ego--90-0.3} \\
    Rotation    & $2$  & \texttt{rotate-z-5-0.5}, \texttt{rotate-z--5-0.5} \\
    Arm raise   & $4$  & \texttt{raisearms-\{l,m\}-\{l,m\}} \\
    Locomotion $+$ arms & $16$ & \texttt{move-arms-$\theta$-$v$-\{l,m\}-\{l,m\}} for $(\theta,v) \in \{(0,0.7),(90,0.7),(180,0.4),(-90,0.7)\}$ \\
    Rotation $+$ arms & $6$  & \texttt{spin-arms-$\pm5$-\{l-l, l-m, m-l\}} \\
    Ground poses & $3$  & \texttt{crouch-0}, \texttt{crouch-0.25}, \texttt{sitonground} \\
    \midrule
    \textbf{Total} & $\mathbf{41}$ & \\
    \bottomrule
  \end{tabular}
\end{table}

\subsection{Marginal transport}
\label{app:transport}
A reward defined on G1 often refers to an absolute configuration that a
differently proportioned target (such as the shorter one T1) cannot reach. When that happens, the there is no pose in the cached buffer for the new robot could return a positive reward. Among these rewards,
the quantity affected by this is \emph{arm-tip height}, which the $26$ arm-related tasks thresholds roughly devided into
a ``low'' band and a ``medium'' band. If those thresholds are directly transfer to other robots, the G1's medium band of
$\geq\!1.0$\,m asks a robot that is only $0.66$\,m tall for something it physically cannot do.

We therefore map each threshold through the two robots' empirical marginals. We compute those
marginals over the $40$ frame-aligned LAFAN retargeted motions, which give us $264{,}625$ frames per robot. We convert each G1 threshold $\theta$ into a quantile of the G1's own arm-tip
height distribution, then use this quantile to trace back the corresponding threshold of the target's distribution, so that
$\theta^{(r)} = \Phi_r^{-1}(\Phi_S(\theta))$. The G1's ``low'' band edges of $0.6$ and $0.8$\,m sit at
quantiles $0.079$ and $0.320$, and its ``medium'' lower edge of $1.0$\,m sits at quantile $0.938$.This way, a robot with a smaller reachable range receives a proportionally tighter ramp instead of an
absolute one it could never satisfy (Table~\ref{tab:bands}).

\begin{table}[h]
  \centering
  \small
  \caption{Arm-height bands after quantile mapping, given as (lower, upper, margin) in metres.}
  \label{tab:bands}
  \begin{tabular}{@{}lccc@{}}
    \toprule
    Robot & Low band & Medium band\\
    \midrule
    M3 & $(0.497,\ 0.824,\ 0.276)$ & $(1.157,\ \infty,\ 0.138)$ \\
    N1 & $(0.416,\ 0.627,\ 0.240)$ & $(0.925,\ \infty,\ 0.120)$ \\
    T1 & $(0.509,\ 0.675,\ 0.172)$ & $(0.820,\ \infty,\ 0.086)$ \\
    \bottomrule
  \end{tabular}
\end{table}

\section{On the Encoder design choice}
\label{app:ablations}

\subsection{Causal encoding for behavior prediction}
\label{sec:exp:causal}
For bidirectional encoders, each frame's latent are passed with
roughly one second of future frames. That is acceptable for offline evaluation, but it requires future observations unavailable during live teleoperation.. We therefore compare bidirectional and causal attention and evaluate their effects on latent alignment and closed-loop tracking (Table~\ref{tab:causal}). We compare the two attention masks at $100\%$ and $25\%$ of the distillation corpus, measured by \emph{alignment} -- the validation cosine against $\latent^\star$ -- and by the closed-loop joint tracking error \emph{joint MAE}.

\begin{table}[h]
  \centering
  \small
  \caption{Causal versus bidirectional attention.}
  \label{tab:causal}
  \begin{tabular}{llcccc}
    \toprule
    Corpus & Encoder & Alignment $\uparrow$ & Joint MAE (rad) $\downarrow$ & $\Delta$ vs.\ oracle \\
    \midrule
    \multirow{2}{*}{$100\%$}
      & bidirectional & $\mathbf{0.8695}$ & $0.2068$ & $+0.0025$ \\
      & causal        & $0.8420$ & $\mathbf{0.2043}$ & $\phantom{+}0.0000$ \\
    \addlinespace
    \multirow{2}{*}{$25\%$}
      & bidirectional & $0.6825$ & $0.2129$ & $+0.0086$ \\
      & causal        & $\mathbf{0.6829}$ & $\mathbf{0.2045}$ & $+0.0002$ \\
    \addlinespace
    ---    & oracle $\latent^\star$ & --- & $0.2043$ & --- \\
    \bottomrule
  \end{tabular}
\end{table}

In both metrics, the causal encoder achieve comparable and performant results compared to its bidirectional counterpart. On the full corpus, causal attention reduces validation cosine similarity by $0.0275$ relative to bidirectional attention, while lowering closed-loop joint MAE from $0.2068$ to $0.2043$ rad, matching the reported MAE under oracle conditioning. With $25\%$ of the corpus, the two encoders achieve nearly identical alignment, while the causal encoder yields lower joint MAE ($0.2045$ versus $0.2129$ rad). These results suggest that, under the evaluated training settings, causal attention maintains tracking performance despite a modest reduction in alignment at full data, while removing the need for future observations and enabling real-time reference streaming.

\subsection{Unified encoder versus embodiment-specific encoders}
\label{app:encoder}

\begin{table}[t]
  \centering
  \small
  \caption{Unified versus per-robot encoders.}
  \label{tab:encoder}
  \begin{tabular}{llccc}
    \toprule
     & & Per-robot & Unified & $\Delta$ \\
    \midrule
    \multirow{3}{*}{Alignment $\uparrow$}
      & M3 & $0.9026 \pm 0.0017$ & $0.8899 \pm 0.0040$ & $-0.0127$ \\
      & T1 & $0.8307 \pm 0.0012$  & $0.8265 \pm 0.0058$ & $-0.0042$ \\
      & N1 & $0.8879 \pm 0.0011$ & $0.8698 \pm 0.0043$ & $-0.0181$ \\
    \addlinespace
    Agreement $\uparrow$
      & all pairs & $0.867$--$0.913$ & $\mathbf{0.938}$--$\mathbf{0.965}$ & $+0.07$ \\
    \bottomrule
  \end{tabular}
\end{table}

Table~\ref{tab:encoder} compares the unified encoder against three independently trained per-robot
encoders. While alignment results (inferred latent cosine distance to ground-truth latent) are similar between per-robot encoders and unified encoder, the unified encoder generates latents with higher agreement among 3 new robots. This illustrates that the unified design significantly 
improves cross-embodiment latent consistency without sacrificing ground-truth alignment. This is because the per-robot encoders are trained on the same data and with the same architecture, but they are not constrained to agree with each other. Together with the generalizability to unseen robots of similar morphology (demonstrated in Section~\ref{sec:exp:loro}), the unified encoder stands as a better choice for a shared behavioral coordinate system enabling commanding the whole platform with the same latent, which is the central goal of this work. We further investigate the effect of the unified encoder's trunk width in Table~\ref{tab:sweep}. Reducing the trunk width from $H{=}512$ to $H{=}256$ \emph{improves} mean cosine by $+0.010$
($0.8548 \to 0.8652$), 
while $H{=}128$ significantly degrades the performance of the encoder.

\begin{table}[t]
  \centering
  \small
  \caption{Trunk width of the unified encoder, measured against $\latent^\star$.}
  \label{tab:sweep}
  \begin{tabular}{lcccc}
    \toprule
    $H$ & M3 & T1 & N1 & Mean \\
    \midrule
    $512$              & $0.8788$ & $0.8254$ & $0.8603$ & $0.8548$ \\
    $256$ (ours)       & $\mathbf{0.8928}$ & $\mathbf{0.8304}$ & $\mathbf{0.8722}$ & $\mathbf{0.8652}$ \\
    $128$              & $0.6968$ & $0.6722$ & $0.6841$ & $0.6844$ \\
    \bottomrule
  \end{tabular}
\end{table}

\subsection{Sensitivity to retargeting error}
\label{app:retarget}
Section~\ref{sec:stage1} relies on the assumption that retargeting already supplies frame-level
correspondence. Here we test sensitivity to violations of this assumption by perturbing the retargeted inputs and their temporal alignment. Corruption is
applied to \textbf{M3 encoder data} while the ground-truth of G1 $\latent^\star$,
the frozen trackers, the reference motions used to train trackers 
are unchanged. We conduct the experiment in two regimes: 1) \emph{training time}: the unified encoder is retrained with 2 uncorrupted dataset and a corrupted one; 2) \emph{inference time}: the trained encoder is fed corrupted input at inference.

\paragraph{Data corruption strategy.} Every corruption is applied on the canonical proprioceptive frames that the encoder takes as input,
rather than the ground-truth labels nor on anything the tracker sees in order to avoid tracker retraining. Two strategies perturb the \emph{geometry} of the correspondence and two perturb its
\emph{timing}:
\begin{itemize}
  \item \textbf{Geometry - Systematic bias ($\times N$).} We draw a fixed per-joint offset vector and apply it
  to every frame of the corpus. This is how a miscalibrated retargeter fails for a wrong joint-limit
  mapping or offset calibration. We
  set $\times1$ to $0.027$\,rad rms, which is the size of the disagreement between two plausible
  tunings of GMR~\citep{ze2026gmr} (the retargeting pipeline used in our paper).
  \item \textbf{Geometry - Non-systematic noise ($\sigma$).} A zero-mean Gaussian perturbation drawn independently
  for each frame and each joint. In practice, this is the more commonly seen mismatch between retargeters, which affects each frame differenly rather than staying as a consistent bias.
  \item \textbf{Timing - Constant timing offset ($\delta$ frames).} Every window is paired with a label $\delta$
  frames away ($\delta{=}1$ equals to $33$\,ms at 30 fps). This error typically belongs to a mismatch in synchronization or motion up/downsampling.
  \item \textbf{Timing - Per-clip timing offset ($k$).} An offset drawn independently for each clip from
  $\mathcal{U}\{-k,\dots,k\}$, which models a retargeter whose per-clip alignment is unreliable.
\end{itemize}

\begin{table}[h]
  \centering
  \small
  \setlength{\tabcolsep}{4pt}
  \caption{\textbf{Deployment regime.} Pretrained encoder with corrupted inference input. \emph{Phase}: per-clip offset that best match a clean clip; \emph{cos@phase}: recovered cosine upon back-shifting.}
  \label{tab:corrupt-deploy}
  \begin{tabular}{lccccc}
    \toprule
    Corruption & Align $\uparrow$ & Phase & cos@phase $\uparrow$ $\uparrow$
      & MAE (rad) $\downarrow$ & Gap closed $\uparrow$ \\
    \midrule
    clean                & $0.8916$ & $0$ & $0.8916$ & $0.1655$ & $99.4\%$ \\
    \addlinespace
    bias $\times1$       & $0.8897$ & $0$ & $0.8897$ & $0.1626$ & $101.1\%$ \\
    bias $\times2$       & $0.8833$ & $0$ & $0.8833$ & $0.1611$ & $101.9\%$ \\
    bias $\times4$       & $0.8580$ & $0$ & $0.8580$ & $0.1647$ & $99.8\%$ \\
    bias $\times8$       & $\mathbf{0.7647}$ & $0$ & $0.7647$ & $\mathbf{0.1913}$ & $84.4\%$ \\
    bias $\times16$      & $\mathbf{0.5210}$ & $0$ & $0.5210$ & $\mathbf{0.2849}$ & $30.2\%$ \\
    Gaussian $\sigma{=}0.05$ & $0.8684$ & $0$ & $0.8684$& $0.1678$ & $98.0\%$ \\
    \addlinespace
    shift $2$ fr         & $0.8623$ & $-2$ & $0.8915$& $0.1664$ & $98.9\%$ \\
    shift $8$ fr         & $0.6589$ & $-8$ & $0.8913$& $0.1890$ & $85.8\%$ \\
    shift $32$ fr        & $0.5416$ & $-32$ & $0.8917$& $0.2127$ & $72.1\%$ \\
    per-clip $k{=}4$     & $0.8692$ & $+1$ & $0.8914$& $0.1721$ & $95.5\%$ \\
    \bottomrule
  \end{tabular}
\end{table}

\begin{table}[h]
  \centering
  \small
  \setlength{\tabcolsep}{4pt}
  \caption{\textbf{Training regime.} The unified encoder retrained with M3 corrupted input.}
  \label{tab:corrupt-train}
  \begin{tabular}{lccccc}
    \toprule
    Corruption & Align $\uparrow$ & Phase & cos@phase $\uparrow$ & MAE (rad) $\downarrow$
      & Gap closed $\uparrow$ \\
    \midrule
    clean                & $0.8632$ & $0$ & $0.8644$ & $0.1669$ & $98.6\%$ \\
    \addlinespace
    bias $\times2$       & $0.8525$ & $0$ & $0.8537$ & $0.1740$ & $94.4\%$ \\
    bias $\times4$       & $0.8242$ & $0$ & $0.8254$ & $0.1848$ & $88.2\%$ \\
    Gaussian $\sigma{=}0.05$ & $0.8427$ & $0$ & $0.8440$ & $0.1675$ & $98.2\%$ \\
    \addlinespace
    shift $2$ fr         & $0.8191$ & $\mathbf{+2}$ & $0.8673$ & $0.1755$ & $93.6\%$ \\
    shift $-2$ fr        & $0.8033$ & $\mathbf{-2}$ & $0.8599$ & $0.1673$ & $98.3\%$ \\
    shift $4$ fr         & $0.7039$ & $\mathbf{+4}$ & $0.8621$ & $0.1853$ & $87.9\%$ \\
    per-clip $k{=}4$     & $0.8191$ & $0$ & $0.8347$ & $0.1734$ & $94.8\%$ \\
    \bottomrule
  \end{tabular}
\end{table}

\begin{table}[h]
  \centering
  \small
  \caption{Data corruption behavior across the two regimes.}
  \label{tab:corrupt-summary}
  \begin{tabular}{lll}
    \toprule
     & Timing ($\delta$, $k$) & Geometry (bias $\times N$, $\sigma$) \\
    \midrule
    Alignment              & collapses ($0.89 \to 0.54$)      & degrades smoothly \\
    Closed-loop tracking   & unchanged once re-phased         & flat to $\times8$, breaks by $\times16$ \\
    Recoverable            & yes, by a one-dimensional search & not shown to be recoverable \\
    \bottomrule
  \end{tabular}
\end{table}

\textbf{Timing: the original timeline can be recovered by a search.} At every magnitude we tried, the
phase scan recovers the injected offset exactly and the cosine returns to $0.8913$--$0.8917$ against a
clean $0.8916$. The latent is not degraded but rather becomes the correct latent for a neighbouring
frame, and even at $32$ frames ($1.07$\,s) retrieval is undisturbed at $0.998$. The tracking performance recovers after the same back-shifting is applied. With a phase search enabled, the corresponding
back-shifting of $8$, $16$ and $32$ frames gives phase-aligned MAE of $0.1646$, $0.1641$ and
$0.1621$\,rad against a clean $0.1655$. The residual $+0.047$ in Table~\ref{tab:corrupt-deploy} is
therefore the cost of executing the right motion at the wrong time, which is a property of the prompt
rather than of the encoder. Training on a shifted corpus has a similar behavior. Here the
encoder \emph{learns} the shifted correspondence rather than correcting it, so when it is fed clean
input, it outputs a latent offset by exactly $\delta$, which is the phase column of
Table~\ref{tab:corrupt-train} that recovers $+2$, $-2$ and $+4$ in both directions. Moreover, the cosine at
that offset returns to $0.8599$--$0.8675$ against a clean $0.8644$. For per-clip random offset, as no global pattern exists, the encoder performs badly even with back-shifting.

\textbf{Geometry: tolerant to non-systematic noise up to $\times4$.} Non-systematic noise by Gaussian perturbation at $\sigma{=}0.05$ costs $+0.002$\,rad during inference and $+0.001$\,rad during training, which is tolerable for our tracking policy. For systematic bias, it is harmless up to roughly $\times4$ and destructive by
$\times8$, demonstrated by the clear drop in both alignment (from $0.8580$ to $0.7647$) and MAE (from $0.1647$ to $0.1913$). We summarize our findings with data corruptions in Table~\ref{tab:corrupt-summary}.

\subsection{Cross-embodiment generalizability experiment protocol}
\label{app:loro}

We compare nine encoders, which are three folds with three seeds. Our training procedure follows Appendix~\ref{app:enc-train}, with modification in the robot set and data loader batch size. As the sampler returns the
same window for every robot, so the effective batch is $\texttt{batch\_size}\times R$. We therefore use
$33$ for the two-robot folds, which gives $33\times2=66$ windows per step and matches the $22\times3$ of
the original encoder. The validation clip split is the quota split of
Section~\ref{sec:exp:generalize}, that is $10$ clips spanning all eight behaviors, and it is the same
for every robot. We measure the random-latent floor inside this harness as well, and it comes out at
$0.3371$ on M3 rather than the $0.4411$ of Section~\ref{sec:exp:scaling} due to the difference in the rollout
protocol (10 clips instead of all $40$ clips). The oracle, predicted and random cells of Table~\ref{tab:loro} all share
one harness and one clip set and are compared amongst each other.

\section{Flow-based Motion Generator}
\label{app:flow}

Because the distilled space $\Zsp$ is shared by every embodiment (Section~\ref{sec:stage1}), a
generator that samples trajectories \emph{inside} $\Zsp$ prompts all robots at once, with no
reference motion and no backward map at inference. We train one conditional rectified-flow
model~\citep{liu2023rectifiedflow,lipman2023flowmatching} over chunks of $H=64$ consecutive latents
$\latent_{1:H}\in\Zsp^{H}$ drawn from the retargeted corpus, analogous to the flow-based action
chunking used in vision-language-action (VLA) models~\citep{li2026vision} or world-action models~\citep{shen2026world, li2026badwam}. The velocity field $v_\theta$ is a
pre-LN transformer~\citep{vaswani2017transformer} ($6$ layers, width $512$, $8$ heads, $\approx16$M
parameters) whose per-frame tokens are $[\latent_t^{(i)}\,\|\,\phi(t)]$ --- the noised latent
concatenated with a sinusoidal embedding of the flow time --- mixed by RoPE self-attention over the
$64$ frames and by cross-attention to the condition tokens described next.

\paragraph{Conditions.} Three conditions enter through two paths. The \emph{mode} prompt --- one
learned embedding per behavior plus a \textsc{null} row, which doubles as the unconditional token
and as the fallback for a clip whose name does not parse (a frozen T5 encoder~\citep{raffel2020t5}
is a drop-in replacement when free-form captions are available) --- and the \emph{history} of the
$16$ preceding latents are both turned into tokens and concatenated into one cross-attention memory,
so every frame can read either. The \emph{motion} condition is instead frame-aligned: the
retargeted features are embedded per frame and added to the corresponding frame token through a
scalar gate initialized to zero, so the generator starts as a pure mode-and-history model and admits
frame-aligned information only as it earns loss. Each condition is dropped independently with
probability $0.1$ during training, which supplies the unconditional branch for guidance and lets any
subset of the three be used at sampling time.

\paragraph{Objective.} For a data chunk $\latent_1$ and noise $\epsilon\sim\mathcal{N}(0,I)$ we take
the straight interpolant $\latent_t=(1-t)\epsilon+t\,\latent_1$ with target velocity
$v^\star=\latent_1-\epsilon$, and minimize the masked conditional flow-matching loss
\begin{equation}
  \mathcal{L}_{\mathrm{FM}}
  = \mathbb{E}_{t,\epsilon,(\latent_1,c)}
    \big\lVert v_\theta(\latent_t,t,c)-v^\star \big\rVert_2^2 ,
  \label{eq:flow}
\end{equation}
averaged over valid frames only, with $t$ drawn from a logit-normal schedule and $c$ the (partially
dropped) conditions. Sampling integrates $\mathrm{d}\latent/\mathrm{d}t=v_\theta$ from $t=0$ to
$t=1$ with $20$ denoising steps and classifier-free guidance~\citep{ho2022cfg} at scale $2.5$,
re-applying the radial projection $\proj$ after every step so the trajectory stays on the sphere of
radius $\sqrt{d}$ that the trackers were trained against. Long sequences are produced
autoregressively: each chunk is conditioned on the last $16$ generated frames. The output is a
latent stream that the frozen latent-conditioned policies of Section~\ref{sec:exp:track} consume
directly --- no tracker, no kinematic decoder, and nothing robot-specific in between.

\paragraph{Data selection and window sampling.} Targets are the latents already baked into the
retargeted corpus by the Stage-1 encoder, so no additional labeling is required. Mode labels are
parsed from the clip name, and a clip that does not parse keeps the \textsc{null} mode rather than
being dropped, so unlabeled data still trains the unconditional branch; clips shorter than the
$64$-frame horizon are discarded. Windows are drawn by sampling a clip uniformly and then a start
index within it, which equalizes exposure across behaviors on a corpus whose clips differ widely in
duration. Evaluation holds out \emph{whole clips}, so the reported flow loss measures
generalization to an unseen motion rather than to an unseen window of a seen motion.

\section{Pipeline Inference}
\label{app:infer}

We extensively evaluate our framework in both simulation and in the real world across all four prompt modes. Full videos in \href{https://dotandung.github.io/crossbfm}{our website}
\begin{figure}[H]
  \centering
  \includegraphics[width=\textwidth,height=0.85\textheight,keepaspectratio]{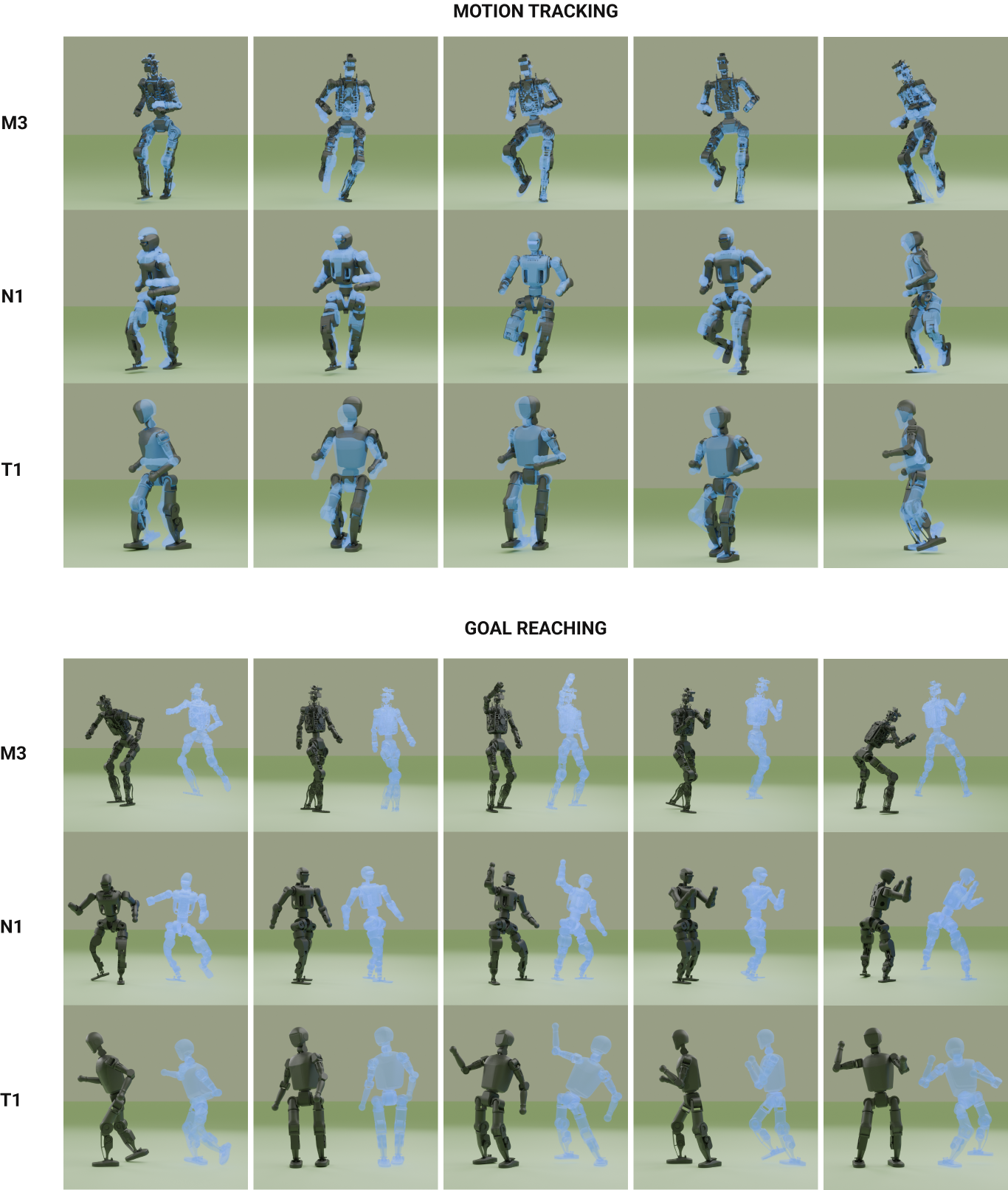}
  \caption{Motion tracking and Goal reaching task for LAFAN motions.}
  \label{fig:dep-tracking}
\end{figure}

\begin{figure}[p]
  \centering
  \includegraphics[width=\textwidth,height=0.92\textheight,keepaspectratio]{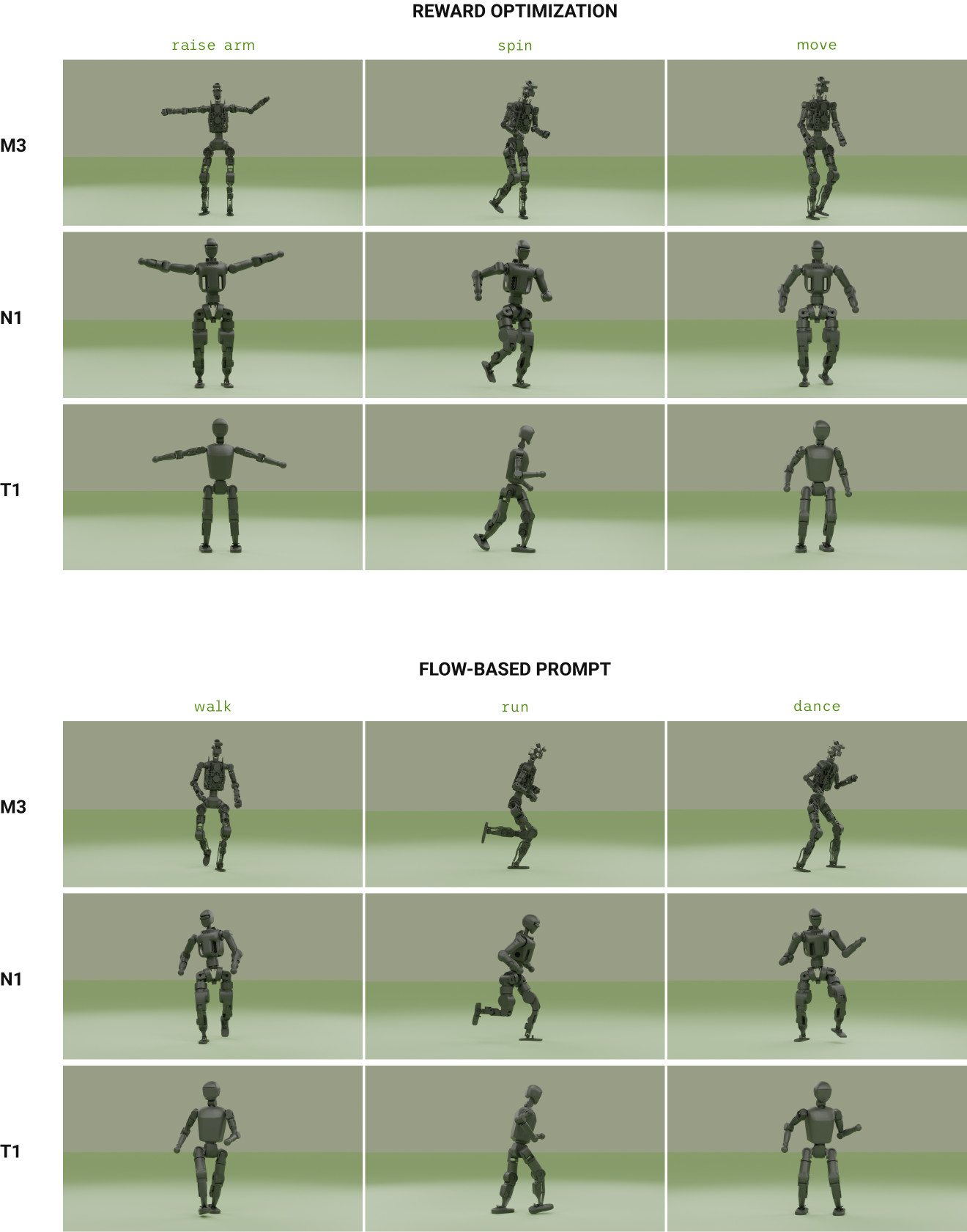}
  \caption{Reward optimization for 3 tasks in the 41-task suite and 3 prompt conditions.}
  \label{fig:dep-reward}
\end{figure}
\clearpage

\end{document}